%% file: main.tex
\documentclass[10pt,twocolumn,letterpaper]{article}

\usepackage{cvpr}              % To produce the CAMERA-READY version
\usepackage{multirow}
\usepackage{booktabs}
\usepackage{adjustbox}
\usepackage{xcolor}
\usepackage{colortbl}
\usepackage{makecell} 
\usepackage[breakable]{tcolorbox}
\usepackage{listings}
\usepackage{tocloft}
\usepackage{float}
\usepackage{tabularx}

\definecolor{cvprblue}{rgb}{0.21,0.49,0.74}
\usepackage[pagebackref,breaklinks,colorlinks,allcolors=cvprblue]{hyperref}
\usepackage{svg}
\def\paperID{xxx} % *** Enter the Paper ID here
\def\confName{CVPR}
\def\confYear{2026}

\title{PhenoLIP: Integrating Phenotype Ontology Knowledge into Medical Vision–Language Pretraining}

\author{
    Cheng Liang$^{1,2}$,\, Chaoyi Wu$^{1}$,\, Weike Zhao$^{1,2}$,\,  \\[4pt]
    Ya Zhang$^{1,2}$,\, Yanfeng Wang$^{1,2}$,\,Weidi Xie$^{1,2}$ \\
    $^{1}$School of Artificial Intelligence, Shanghai Jiao Tong University \\
    $^{2}$Shanghai Artificial Intelligence Laboratory\\[2pt]
    \url{https://github.com/MAGIC-AI4Med/PhenoLIP}
}

\begin{document}
\twocolumn[{%
\renewcommand\twocolumn[1][]{#1}%
\maketitle
\vspace{-30pt}
\begin{center}
   \centering
   \includegraphics[width=\textwidth]{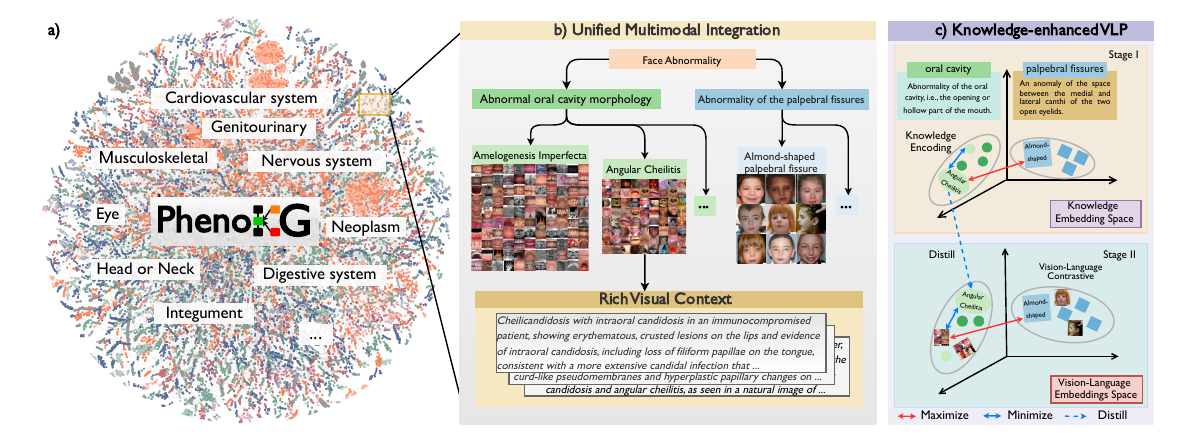}
   \vspace{-15pt}
   \captionof{figure}{
        The figure illustrates our core methods \textbf{PhenoKG} and \textbf{PhenoLIP}, from left to right: \textbf{(a)} macro-level visualization of PhenoKG, the first large-scale phenotype-centric multimodal knowledge graph that hierarchically organizes diverse anatomical systems; \textbf{(b)} Unified multimodal integration that aligns phenotype images, rich visual descriptions, and structured phenotype ontology knowledge into a single graph; and \textbf{(c)} \textbf{PhenoLIP}, our phenotype knowledge-enhanced vision–language pretraining framework, which first learns a structured phenotype embedding from the ontology and then guide the vision-language pretraining via distillation.
   }
  \label{fig:overview}
 \end{center}
}]
% \maketitle
% \vspace{-30pt}
% % Teaser figure - 简单直接的方式
% \begin{figure*}[t]
%   \centering
%   \includegraphics[width=\textwidth]{figs/overview.pdf}
%   \caption{
%     The figure illustrates our core methods \textbf{PhenoKG} and \textbf{PhenoLIP}, from left to right: \textbf{(a)} macro-level visualization of PhenoKG, the first large-scale phenotype-centric multimodal knowledge graph that hierarchically organizes diverse anatomical systems; \textbf{(b)} Unified multimodal integration that aligns phenotype images, rich visual descriptions, and structured phenotype ontology knowledge into a single graph; and \textbf{(c)} \textbf{PhenoLIP}, our phenotype knowledge-enhanced vision–language pretraining framework, which first learns a structured phenotype embedding from the ontology and then guide the vision-language pretraining via distillation.
%   }
%   \label{fig:overview}
% \end{figure*}

\input{sec/0_abstract}    
\input{sec/1_intro}
\input{sec/2_related_works}

\input{sec/3_methods}

\input{sec/4_experiments}

\input{sec/5_conclusion}

\clearpage

\twocolumn
{
    \small
    \bibliographystyle{ieeenat_fullname}
    \bibliography{main}
}

% 附录
\clearpage
\input{sec/X_suppl.tex}

\end{document}

%% file: sec/0_abstract.tex
\begin{abstract}
Recent progress in large-scale CLIP-like vision-language models~(VLMs) has greatly advanced medical image analysis. 
However, most existing medical VLMs still rely on coarse image-text contrastive objectives and fail to capture the systematic visual knowledge encoded in well-defined medical phenotype ontologies. 
To address this gap, we construct \textbf{PhenoKG}, the first large-scale, phenotype-centric multimodal knowledge graph that encompasses over 520K high-quality image-text pairs linked to more than 3,000 phenotypes. 
Building upon PhenoKG, we propose \textbf{PhenoLIP}, a novel pretraining framework that explicitly incorporates structured phenotype knowledge into medical VLMs through a two-stage process. 
We first learn a knowledge-enhanced phenotype embedding space from textual ontology data and then distill this structured knowledge into multimodal pretraining via a teacher-guided knowledge distillation objective. 
To support evaluation, we further introduce \textbf{PhenoBench}, an expert-verified benchmark designed for phenotype recognition, comprising over 7,800 image--caption pairs covering more than 1,000 phenotypes.
Extensive experiments demonstrate that PhenoLIP outperforms previous state-of-the-art baselines, improving upon BiomedCLIP in phenotype classification accuracy by 8.85\% and BIOMEDICA in cross-modal retrieval by 15.03\%, underscoring the value of integrating phenotype-centric priors into medical VLMs for structured and interpretable medical image understanding.

\end{abstract}

%% file: sec/1_intro.tex
\vspace{-4mm} 
\section{Introduction}
\label{sec:intro}

Large-scale vision-language model pretraining, such as CLIP~\cite{clip}, has made significant progress in computer vision, which has also inspired medical image analysis. 
Current medical VLMs~\cite{pmc-clip,biomedica,biomedclip}, trained on massive collections of medical image--text pairs through contrastive learning, demonstrate strong cross-modal understanding and generalization capabilities. 
They have revolutionized a wide range of medical visual and multimodal tasks, \emph{e.g.}, disease diagnosis~\cite{pmc-clip,llavamed, qiu2025evolving}, radiology report generation~\cite{radfm,chexpert}, and medical visual question answering~\cite{chestxreasoner}, achieving expert-level performance on several benchmarks. 

Despite these impressive achievements, current medical VLMs largely rely on data-driven associations, lack systematic integration of structured medical knowledge and accumulated expertise that guides clinical practice and research across multiple specialties. 
In medical image analysis, one of the most fundamental and essential ontology systems is the \textbf{phenotype ontology}, which abstracts diverse medical image patterns and textual clinical descriptions into hierarchical phenotype terms that are further linked to corresponding diseases. 
Such structured knowledge supports recognizing fine-grained visual distinctions in medical images, perform cross-case comparisons, and associate visual patterns with diagnostic knowledge. 
Therefore, integrating phenotype ontology knowledge into the pretraining process can greatly enhance medical VLMs’ ability to comprehensively understand medical image semantics.

In this paper, we construct the first, large-scale, phenotype-centric multimodal medical knowledge graph, termed as \textbf{PhenoKG}, providing the structured and multimodal knowledge that current VLMs lack. Our approach builds upon the Human Phenotype Ontology (HPO)~\cite{hpo}, a well-established taxonomy of human phenotypes with rich clinical annotations and disease associations. To incorporate multimodal information, we develop a scalable pipeline that automatically mines and filters medical image-caption pairs from literature. The resulting high-quality image–caption pairs are linked to phenotype nodes, enriching the knowledge graph with multimodal evidence. Unlike prior vision–language datasets built from unstructured web data~\cite{laion,biomedica}, \textbf{PhenoKG} unifies visual data with structured phenotypic knowledge, bridging medical image understanding with ontology-based priors. In total, \textbf{PhenoKG} comprises 5{,}839 phenotype groups and 13{,}812 specific phenotypes, including 524{,}804 image–caption pairs aligned with 3{,}096 phenotypes, establishing a systematic foundation for medical Vision-Language Pretraining.

Departing from conventional medical VLMs such as BiomedCLIP~\cite{biomedica} and BIOMEDICA~\cite{biomedica}, which are typically pretrained on large but unstructured collections of image-caption pairs, our approach is fundamentally knowledge-driven. We propose \textbf{P}henotype-centric \textbf{L}anguage--\textbf{I}mage \textbf{P}retraining, termed as \textbf{PhenoLIP}, a two-stage pretraining framework designed to gradually distill the structured multimodal medical knowledge into VLMs. 
%Specifically, we propose a two-stage procedure that gradually incorporates knowledge from textual to multimodal sources.  
We first train a phenotype knowledge encoder on the textual ontology within \textbf{PhenoKG} to map phenotype names, definitions, and relationships into a unified semantic space. Optimized with contrastive learning, the encoder pulls semantically related phrases, such as synonyms, definitions, and hierarchical relations, closer in the embedding space. 
In the second stage, during the main vision-language pretraining on the multimodal data within PhenoKG, we employ the frozen knowledge encoder as a teacher to guide the VLM's text encoder.  Through an auxiliary knowledge distillation objective, the text encoder is regularized to align its outputs with the teacher's knowledge-enhanced embedding space. This process runs in parallel with the primary image-text contrastive alignment, effectively injecting phenotype ontology priors into the VLM and enabling it to ground visual evidence in fine-grained, structured phenotype concepts.

For evaluation, we introduce \textbf{PhenoBench}, an expert-verified benchmark specifically designed for phenotype recognition, that contains 7,819 image-caption pairs covering 1,187 distinct phenotypes.
This task represents a fundamental capability in medical image analysis that underpins more advanced applications, such as diagnosis, medical visual question answering, and report generation.

We evaluate our final pre-trained VLMs on both PhenoBench and multiple public datasets across diverse tasks, including {phenotype classification}, {fine-grained image-to-phenotype retrieval}, and {rare facial phenotype identification}. Our method outperforms state-of-the-art biomedical VLM by 8.85\% average accuracy in phenotype classification and 15.03\% in cross-modal retrieval on 9 downstream datasets, including our proposed \textbf{PhenoBench}. The results underscore the effectiveness of our approach and highlight the transformative potential of integrating structured knowledge with multimodal pre-training to enable more precise and semantically-grounded reasoning in medical AI.

%% file: sec/2_related_works.tex
\section{Related Work}
\label{sec:related_works}

\noindent \textbf{VLP in medical image analysis.}

Vision–Language Pre-training (VLP) aims to learn joint representations from large-scale image–text data. General-domain models such as CLIP~\cite{clip} and its successors~\cite{siglip2, coca} have achieved strong performance by aligning images and text in a shared embedding space using contrastive learning. However, when transferred to specialized domains like medicine, these models suffer from a vocabulary gap and limited ability to capture fine-grained clinical semantics~\cite{laion}, underscoring the need for domain-specific pretraining.
Recent medical VLMs have extended CLIP-style frameworks into biomedicine. BiomedCLIP~\cite{biomedclip} pre-trains on 15M biomedical image-text pairs from PubMed, while PMC-CLIP~\cite{pmc-clip} leverages figure-caption pairs from PubMed Central to learn general biomedical representations.

{
\vspace{3pt}\noindent \textbf{Knowledge-enhanced medical VLMs.}
Beyond general medical VLMs, several approaches have explored incorporating structured medical knowledge into vision-language models. DermLIP~\cite{Derm1M} constructs a dermatology-specific dataset and employs ontology-based text refinement to enhance representations. KEP~\cite{kep} focuses on pathology images and incorporates structured disease knowledge to enhance vision-language alignment. 
In the radiology domain, MedKLIP~\cite{medklip} extracts report information into structured triplets and enriches entities with knowledge-based descriptions.
KAD~\cite{kad} leverages a medical knowledge graph (UMLS~\cite{umls}) represented as concept–relation–concept triplets, and uses entities/relations extracted from reports to guide vision–language pretraining.
Although existing knowledge-enhanced approaches excel in specialized domains, they neglect the critical role of \textit{phenotype ontology knowledge}. This ontology defines atom-level, fine-grained pathology priors, providing a comprehensive knowledge foundation for medical imaging tasks across all modalities.
}

\vspace{3pt}\noindent \textbf{Medical knowledge graph.}
Medical Knowledge Graphs (KGs) are essential for organizing complex biomedical information. Large-scale KGs like UMLS~\cite{umls}, OrphaNet~\cite{orphanet}, and OMIM~\cite{omim} have structured vast amounts of data on terminologies, rare diseases, and genes. iKraph~\cite{ikraph}, PrimgKG~\cite{primekg} links molecular and genetic factors to clinical outcomes by structuring 4 million relationships across entities like proteins, biological processes, phenotypes, and drugs.
Among these, the Human Phenotype Ontology (HPO)~\cite{hpo} provides a standardized, hierarchical vocabulary for human phenotypic abnormalities. However, existing medical KGs, including HPO, are predominantly text-based and lack explicit links to visual data, which severely restricts their utility in medical image analysis. Our work aims to bridge this modality gap.

\vspace{3pt}\noindent \textbf{Medical vision-language datasets.}
In the recent literature, significant progress has been made for curating large-scale, high-quality image–caption data in medical domains. 
BIOMEDICA~\cite{biomedica}, PMC-CLIP~\cite{pmc-clip}, and PMC-15M~\cite{biomedclip} constructed datasets covering diverse medical images, charts, and related text descriptions through automated crawling strategies. 
Derm1M~\cite{Derm1M} focused on dermatology, providing rich image–caption pairs through fine-grained annotation and classification. 
Quilt-1M~\cite{quilt1m} collected pathology-related videos and their subtitles from YouTube, generating paired image–text data through multi-stage preprocessing. 
Although these datasets have greatly expanded medical multimodal resources, they remain fragmented and lack a systematic organization aligned with phenotype ontology systems. 
This absence of structured phenotype-centered knowledge limits their effectiveness for developing comprehensive medical VLMs.

%% file: sec/3_methods.tex
\begin{figure*}[h]
    \centering
    \includegraphics[width=\linewidth]{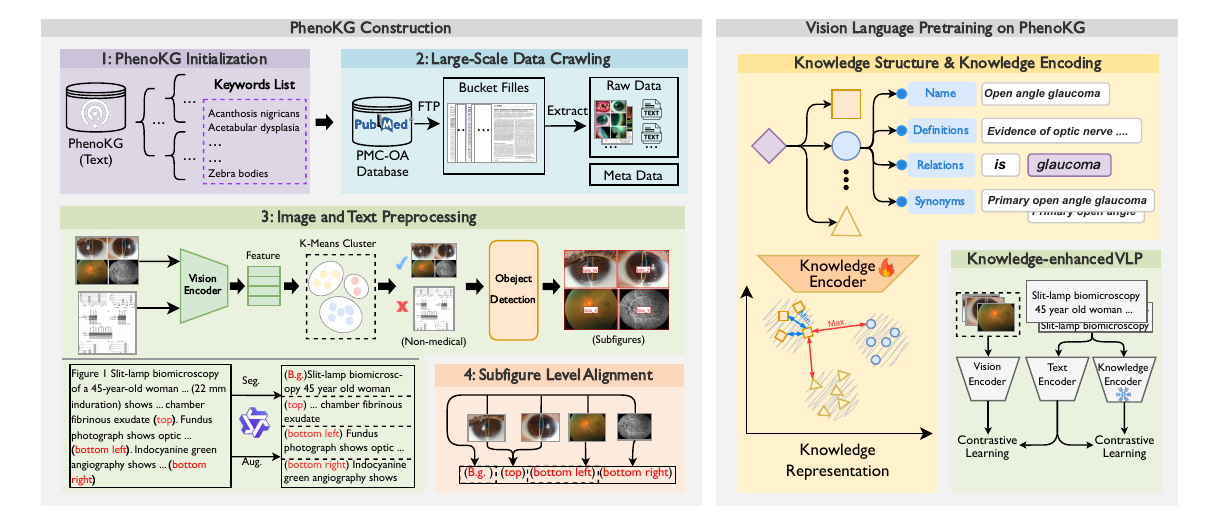}
    \caption{Overview of the \textbf{PhenoKG} construction pipeline and the \textbf{PhenoLIP} training process. \textbf{Left}: PhenoKG Construction. This illustrates the four-stage pipeline for building our multimodal knowledge graph: (1) PhenoKG Initialization, (2) Crawling image-caption pairs from the PMC-OA database using phenotype keywords, (3) Preprocessing images (via clustering-based filtering and subfigure detection) and text (via LLM-based refinement), (4) Aligning subfigures with their corresponding descriptions. \textbf{Right}: PhenoLIP Pretraining. This depicts our two-stage pretraining framework. First, a knowledge encoder is trained on PhenoKG's textual ontology to learn a structured phenotype embedding. Second, during vision-language pretraining, this frozen knowledge encoder serves as a teacher to distill its structured knowledge into the VLM's text encoder, complementing the primary image-text contrastive alignment.}
    \label{fig:pipeline}
    \vspace{-13pt}
\end{figure*}

\section{PhenoKG \& PhenoBench Construction}
\label{sec:phenoKG}

In this section, we detail the construction of a large-scale multimodal phenotype knowledge graph. As illustrated in Figure~\ref{fig:pipeline} (left), the procedure contains two main stages. First, we initialize its textual part based on the Human Phenotype Ontology~(HPO) (Sec.~\ref{sec:keyword_collection}). Second, we develop an automated pipeline to augment this graph with over 524,804 aligned image-caption pairs obtained from biomedical literature (Sec.~\ref{sec:phenopmc_collection}), the detailed statistics are listed in Appendix~\ref{suppsec:phenokg_stat}. Finally, by splitting the crawled articles, we also construct \textbf{PhenoBench}, an expert-annotated benchmark for phenotype recognition evaluation.

\subsection{Knowledge Graph Initialization}
\label{sec:keyword_collection}
To establish \textbf{PhenoKG}, we first construct its textual part based on the Human Phenotype Ontology (HPO)~\cite{hpo}, a standardized ontology system for phenotypes.
We collected 19{,}703 phenotype terms as graph nodes, together with their definitions, relational edges, and synonyms from the Human Phenotype Ontology (HPO). 
This collection serves as an initialized text-only phenotype knowledge graph, comprising 19{,}703 nodes and 23{,}528 edges that represent hierarchical relationships and correlations among phenotype terms. The construction details and statistics of this process are provided in Appendix~\ref{suppsec:kg_construction_text}.

\subsection{Multimodal Data Collection and Processing}
\label{sec:phenopmc_collection}

After constructing the textual phenotype KG, we enriched each phenotype node with multimodal evidence from the biomedical literature using the following pipeline.

\vspace{2pt} \noindent \textbf{Data crawling.} 
Based on the constructed knowledge graph, we first filter a list of keywords. 
We retained only terminal nodes, \textit{i.e.}, those without outgoing edges, and removed other non-terminal, categorical nodes such as ``abnormality of the nervous system''.
Leveraging this keyword list, we processed the entire PubMed Central Open Access (PMC-OA) corpus~\cite{pmcoa}, extracting figure–caption pairs whose captions contained exact matches to any phenotype keyword. In addition to captions, we retrieved the corresponding in-text figure-referencing paragraph (e.g., “...as seen in Figure~1...”), preserving contextual cues for later disambiguation and grounding.

\vspace{2pt} \noindent \textbf{Image processing.}
As illustrated in Figure~\ref{fig:pipeline}, raw figure–caption pairs contain substantial noise, including compound figures and non-medical graphics ({\em e.g.}, charts and graphs). 
First, to remove non-medical images, we embedded all figures with DINOv3 and clustered embeddings via K-means (k = 400)~\cite{scikit, dinov3}. We then screened clusters to retain medically relevant imagery. Second, to split compound figures, we applied a pretrained subfigure detector (DAB-DETR)~\cite{dabdetr,openpmc-18m} to segment multi-panel figures into subfigures. Third, we re-applied the clustering-based filter to the extracted subfigures to eliminate residual non-medical content. This procedure yielded 524{,}804 medical figure–caption pairs. Additional implementation details are provided in Appendix~\ref{sec:image_processing}.

\vspace{2pt} \noindent \textbf{Text processing.} 
\label{para:txt_processiing}
For each retained figure, we kept the original caption and the associated in-text paragraph. 
We then used Qwen3~\cite{qwen3} to synthesize a cleaned caption per image, removing non-informative tokens and numeric artifacts while preserving semantic content. For compound figures that were segmented into subfigures, we further prompted the LLM with a tailored instruction to produce subfigure-level textual descriptions consistent with the original semantics~(Appendix~\ref{sec:text_splitting}).

\vspace{2pt} \noindent \textbf{Subfigure level alignment.}
We aligned subfigures with their corresponding subfigure texts. If the number of subfigures and candidate texts did not match, we conservatively retained the original compound figure–caption pair. Otherwise, we invoked Qwen2.5-VL~\cite{qwen2.5vl} to perform fine-grained matching. Specifically, we rendered the compound image overlaid with subfigure bounding boxes and unique identifiers~({\em e.g.}, \textit{box\_1}, \textit{box\_2}) and prompted the model to map each identifier to the most relevant text segment (Fig.~\ref{fig:pipeline}; Appendix~\ref{suppsec:subfigure_processing}). An ablation on overlaying boxes for prompting is reported in Appendix~\ref{sec:subfigure_detection}.

\vspace{2pt} \noindent \textbf{Multimodal integration.}
The final step was to integrate the collected image-caption pairs into the original textual phenotype KG nodes. Since each image–caption pair is filtered by search keywords that are directly linked to the original phenotype nodes, it is straightforward to insert them into the textual KG structure based on keywords.

\subsection{PhenoBench Construction.}
In parallel with PhenoKG, we curated an independent evaluation suite, \textbf{PhenoBench}, using a matched but disjoint pipeline to prevent data leakage.
We enforced a strict split at the document and figure levels by partitioning the source corpus using unique \texttt{PMCID} and \texttt{FigureID}. Articles and figures assigned to PhenoBench are excluded from any training or development sets, ensuring no overlap of images, captions, or near-duplicates across splits.
All candidate image–caption pairs were manually reviewed by clinical experts to confirm medical relevance, caption fidelity to the visual content, and accurate phenotype linkage. 
The final benchmark contains 7,819 image–caption pairs linked to 1,187 distinct phenotype terms and can be used to evaluate phenotype recognition (based on the linked phenotype terms) and image–text retrieval (based on the original image–caption pair supervision).

\section{Vision-Language Pretraining on PhenoKG}
\label{sec:knowledge_pretraining}
We present \textbf{PhenoLIP}, a vision–language pretraining framework that distills structured phenotype knowledge from PhenoKG into joint visual–textual representations (Figure~\ref{fig:pipeline}). The training comprises two stages. First, we train a phenotype-aware text encoder to capture relational structure over the phenotype KG (Sec.~\ref{sec:knowledge_encoding}). Second, we use this knowledge-enhanced encoder to guide vision–language contrastive pretraining, aligning images with text while injecting KG priors into the VLM (Sec.~\ref{sec:vl_align}).

\subsection{Phenotype Knowledge Encoding}
\label{sec:knowledge_encoding}

To leverage the phenotype knowledge in PhenoKG, we learn text embeddings for phenotype concepts in a space explicitly regularized by ontology relations, hierarchical structure, and lexical semantics. As illustrated in Figure~\ref{fig:pipeline}~(Right), we train a phenotype-aware text encoder with metric learning to map each terms, their definitions, synonyms, and hierarchical (is‑a) relations into a latent space, enabling effective utilization of phenotype knowledge. 

\vspace{3pt} \noindent \textbf{Problem formulation.}
Given a set of phenotype nodes, each can be denoted as $(T_i, A_i)$, where $T_i$ represents the phenotype term, and $A_i$ represents its related attribute set, including various textual types: ({i}) the phenotype name, ({ii}) its detailed definition, ({iii}) its synonymous names, and ({iv}) relational descriptions derived from its related hierarchical (is-a) relations (e.g., ``\textit{[head phenotype] is a child phenotype of [tail phenotype]}'').
Our goal is to learn a knowledge encoder \( \Phi_{\text{k}}(\cdot) \) that satisfies:
\begin{align}
    \text{sim}(\Phi_{\text{k}}(a_i), \Phi_{\text{k}}(a_i^+)) \gg 
    \text{sim}(\Phi_{\text{k}}(a_i), \Phi_{\text{k}}(a_j)), \text{ } i \neq j,
    \label{eq:0}
\end{align}
where $\text{sim}(\cdot, \cdot)$ denotes the similarity function, $a_i$ and $a_i^+$ represent two attributes sampled from the phenotype attribute set $A_i$, and $a_j$ is an attribute sampled from another phenotype set $A_j$.
The encoder can map related phenotype attributes to similar vector representations, while mapping unrelated concepts to distant vector representations.

\vspace{3pt} \noindent \textbf{Training pipeline.}
To optimize the objective in Eq.~\ref{eq:0}, we train the knowledge encoder with a contrastive objective that pulls together multiple textual descriptions of the same phenotype concept and pushes apart descriptions from different concepts.
Specifically, given a sampled attribute $a_i$, we first sample another positive attribute pair $a_i^+$ from PhenoKG and encode them into textual embeddings with the knowledge encoder, termed as $z=\Phi_\text{k}(a)$, resulting in $(z_i, z_i^+)$. 
Given a mini training batch with $2B$ batch size, which consists of \(B\) phenotypes, each with two randomly sampled attribute descriptions, we adopt an InfoNCE-style loss with in-batch negatives:
\begin{align}
    \mathcal{L}_\text{k}= - \frac{1}{2B}\sum_{i=1}^{2B} \log \frac{\exp(\text{sim}(z_i, z_i^+)/\tau_1)}{\sum_{k=1}^{2B} \mathbf{1}_{k \neq i} \exp(\text{sim}(z_i, z_k)/\tau_1)},
\end{align}
where $\text{sim}(\cdot,\cdot)$ denotes cosine similarity, $\tau_1$ is a temperature hyperparameter. By minimizing $\mathcal{L}_{\text{k}}$, the knowledge encoder $\Phi_{\text{k}}$ learns to project semantically related textual attribute descriptions of the same phenotype terms more closely in the embedding space.

\vspace{3pt} \noindent \textbf{Implementation details:}
We adopt the pretrained PubmedBERT~\cite{pubmedbert} as the initialization of the knowledge encoder. More details is listed in Appendix~\ref{suppsec:kg_encoding}.

\subsection{Knowledge-Enhanced VLP}
\label{sec:vl_align}

This section describes the details of knowledge-enhanced visual representation learning, leveraging the former phenotype knowledge encoder $\Phi_\text{k}$ trained on PhenoKG. 

\vspace{3pt} \noindent \textbf{Problem formulation.} 
Given the image--caption pairs collected from PhenoKG, denoted as $(x_i, c_i)$, 
our objective is to train a visual encoder $\Phi_{\text{v}}$ and a text encoder $\Phi_{\text{t}}$, such that their joint embedding space satisfies:
\begin{align}
    \text{sim}(\Phi_{\text{v}}(x_i), \Phi_{\text{t}}(c_i)) \gg \text{sim}(\Phi_{\text{v}}(x_i), \Phi_{\text{t}}(c_j)), \text{ } i \neq j.
    \label{eq:vlp_objective}
\end{align}

\vspace{3pt} \noindent \textbf{Training pipeline.} 
During training, as illustrated in Figure~\ref{fig:pipeline}, we adopt InfoNCE-style contrastive learning to align visual and textual representations. In addition, to ensure that the text embedding space effectively captures the hierarchical knowledge encoded in the knowledge graph (KG), we introduce an auxiliary knowledge distillation objective, which regularizes the learned text embeddings to align with the pretrained knowledge encoder.

Specifically, denoting the visual embedding as $v_i=\Phi_\text{v}(x_i)$ and text embedding as $t_i=\Phi_\text{t}(c_i)$, the multimodal contrastive learning loss can be defined as:
\begin{align}
\mathcal{L}_{\text{M}}
= -\frac{1}{B} \sum_{i=1}^{B} \!
\Bigg[
& \log \frac{\exp(\text{sim}(v_i, t_i)/\tau_2)}
{\sum_{j=1}^{B} \exp(\text{sim}(v_i, t_j)/\tau_2)} \nonumber \\
& + \log \frac{\exp(\text{sim}(t_i, v_i)/\tau_2)}
{\sum_{j=1}^{B} \exp(\text{sim}(t_i, v_j)/\tau_2)}
\Bigg],
\end{align}
where $\tau_2$ is a temperature parameter and $B$ is the training batch size. 
Additionally, we introduce a knowledge distillation branch with the pretrained knowledge encoder $\Phi_{\text{k}}$ acting as a teacher model. For each caption $c_i$ in the batch, the teacher produces a reference knowledge embedding $\mathbf{k}_i = \Phi_{\text{k}}(c_i)$. We then enforce consistency between the student's text embedding $\mathbf{t}_i$ and the teacher's knowledge embedding $\mathbf{k}_i$ using a text-knowledge contrastive loss, denoted as $\mathcal{L}_{\text{KD}}$. This loss function encourages the student's text encoder to map captions into the pre-structured semantic space of the teacher by treating $(\mathbf{t}_i, \mathbf{k}_i)$ as a positive pair and all other non-corresponding pairs $(\mathbf{t}_i, \mathbf{k}_j)$ as negatives within the batch. The detailed formulation of $\mathcal{L}_{\text{KD}}$ is provided in Appendix~\ref{app:kd_loss}.

\vspace{3pt} \noindent \textbf{Implementation details.}
We adopt BiomedCLIP~\cite{biomedclip} as the initialization for the visual encoder, while the text encoder is initialized with the weights from the trained phenotype knowledge encoder. Details are in Appendix~\ref{suppsec:ke_vlp}.

%% file: sec/4_experiments.tex
    \section{Experiments}
    \label{sec:experiments}

    In this section, we introduce our experiments in detail, 
    including evaluation settings, related datasets, and metrics.

    \vspace{3pt} \noindent \textbf{Evaluation settings.} 
    We assess three standard regimes: zero-shot classification, cross-modal retrieval, and linear probing. In \textbf{zero-shot classification}, the model assigns labels to unseen tasks by comparing image embeddings with class-specific textual prompts, without fine-tuning, following CLIP~\cite{clip}; prompt templates are in Appendix~\ref{suppsec:evaluation_prompts}. 
    In \textbf{cross-modal retrieval}, we evaluate image-to-text and text-to-image retrieval to measure alignment between image and phenotype descriptions.
    In \textbf{linear probing}, we freeze the pretrained encoders and train a single linear layer on top of the visual features to gauge representation quality.

    % cls
    \begin{table*}[t] 
        \footnotesize 
        \centering 
        \setlength{\tabcolsep}{2.2pt} 
        \begin{tabular}{lcccccccccc} 
        \toprule 
        \multirow{2}{*}{\textbf{Method}} & \multicolumn{2}{c}{\textbf{Encoder}} & \multicolumn{7}{c}{\textbf{Accuracy}} \\ 
        \cmidrule(lr){2-3} \cmidrule(lr){4-10}
        & \textbf{Vision} & \textbf{Text} & \textbf{Dermatology} & \textbf{Pathology} & \textbf{Radiology} & \textbf{Hematology} & \textbf{Histology} & \textbf{Ophthalmology} & \textbf{PhenoBench} & \textbf{Average} \\ 
        \midrule 
        \multicolumn{11}{c}{\cellcolor{gray!15}\textbf{\textit{General VLMs}}} \\
        \midrule
        OpenCLIP~\cite{openclip} & ViT-B & GPT2 & 14.77 & {33.33} & {29.75} & \underline{23.00} & 4.52 & 19.43 & {2.26} & {18.15} \\ 
        SigLIP2~\cite{siglip2} & So400m & SigLIP64 & 11.31 & 20.00 & 4.95 & 8.30 & \textbf{9.00} & 23.00 & 0.25 & 10.97 \\ 
        CoCa~\cite{coca} & ViT-B & GPT2 & 12.63 & 33.00 & 25.87 & 6.55 & 3.39 & 20.55 & 1.38 & 14.77 \\ 
        \midrule 
        \multicolumn{11}{c}{\cellcolor{gray!15}\textbf{\textit{Biomedical VLMs}}} \\
        \midrule
        PMC-CLIP~\cite{pmc-clip} & ResNet50 & PubmedBert & 41.35 & {45.12} & 29.10 & 21.50 & 5.68 & \underline{43.75} & 7.50 & {27.71} \\
        BiomedCLIP~\cite{biomedclip} & ViT-B & PubmedBert & {47.59} & {42.87} & 28.47 & {20.40} & 4.23 & {40.18} & \underline{8.15} & {27.41} \\
        BIOMEDICA~\cite{biomedica} & ViT-L & GPT2 & \textbf{56.76} & \underline{57.40} & \underline{37.72} & 9.35 & {5.40} & 26.15 & 6.69 & \underline{28.50} \\ 
        \midrule
        \rowcolor{green!15}
        \textbf{PhenoLIP} & ViT-B & PubmedBert & \underline{55.08} & \textbf{58.33} & \textbf{49.03} & \textbf{23.24} & \underline{7.87} & \textbf{51.63} & \textbf{10.76} & \textbf{36.56} \\ 
        \bottomrule 
        \end{tabular} 
        \vspace{-2mm} 
        \caption{Benchmarking results on zero-shot image classification (Acc). Best method per column is \textbf{bold}; second-best model is \underline{underlined}.} 
        \vspace{-2mm}
        \label{tab:zs} 
    \end{table*}

    \vspace{3pt} \noindent \textbf{Evaluation datasets.} 
    We benchmarked on diverse biomedical datasets spanning 7 imaging modalities and 10 medical image analysis tasks under the three regimes above. Dataset descriptions, splits, evaluation protocols, and their clinical features are detailed in Appendix~\ref{suppsec:evaluation_prompts}.

    \begin{itemize}
    \setlength\itemsep{3pt}

        \item For zero-shot classification, we employ {HAM10000}~\cite{ham10000} and {DermaMNIST}~\cite{medmnistv2} for dermatology, {LC25000}~\cite{lc25000} for pathology, {BreastMNIST}, {ChestMNIST}, and {RSNA}~\cite{cxr8} for radiology, and {BloodMNIST}, {TissueMNIST}, {RetinaMNIST}, and {OCTMNIST}~\cite{medmnistv2} for hematology, histology, and ophthalmology, respecively, along with our proposed {PhenoBench} for phenotype recognition. 
        
        \item For cross-modal retrieval, we adopt {PhenoBench} across four distinct retrieval scenarios: I2T~(image-to-text), T2I, I2P~(image-to-phenotype) and P2I retrieval. Additionally, we involve {Face2Gene}~\cite{face2gene}, to evaluate the model's performance in rare disease related facial phenotype retrieval, denoting a fine-grained and long-tail benchmark that challenges the model to associate subtle visual traits with corresponding phenotype concepts.
        
        \item For linear probing, we adopt seven benchmark datasets, including RSNA~\cite{cxr8}, HAM10000~\cite{ham10000}, and five datasets from MedMNIST~v2~\cite{medmnistv2}, including BreastMNIST, ChestMNIST, DermaMNIST, OCTMNIST, and RetinaMNIST. We report the results under 1\%, 10\%, and 100\% of the training data, respectively.
    \end{itemize}

    \vspace{3pt} \noindent \textbf{Baselines.}
    We consider two main categories of state-of-the-art vision–language models (VLMs) for comparison, \emph{i.e.}, {general-domain VLMs} and {biomedical VLMs}.
    The former category includes Open-CLIP~\cite{openclip}, SigLIP2~\cite{siglip2}, and CoCa~\cite{coca}, which are pre-trained on large-scale general-domain web image–text data.
    The latter category comprises PMC-CLIP~\cite{pmc-clip}, BiomedCLIP~\cite{biomedclip}, and BIOMEDICA~\cite{biomedica}, all pre-trained on domain-specific medical image–text data.
    {
    We also compare with knowledge-enhanced medical VLMs, including DermLIP~\cite{Derm1M}, KEEP~\cite{keep}, MedKLIP~\cite{medklip} and KAD~\cite{kad}, which inject medical knowledge through different mechanisms (details in Appendix~\ref{suppsec:ke_baselines}).
    A comprehensive cross-domain performance comparison with these knowledge-enhanced baselines is provided in Appendix~\ref{suppsec:ke_results}.
    }

    \vspace{3pt} \noindent \textbf{Metrics.}
    Model performance is evaluated with metrics tailored to each setting. For zero-shot classification and linear probing, we report {classification accuracy (ACC)} to measure recognition and representation quality. For cross-modal retrieval tasks, including the rare facial phenotype matching on {Face2Gene}~\cite{face2gene}, we use {Recall@K (R@K)} as the main evaluation metric, along with {Precision}, {Recall}, and {F1-Score} to capture fine-grained matching performance in clinically relevant retrieval scenarios.

    % retrieval
    \begin{table*}[t]
        \footnotesize
        \centering
        \setlength{\tabcolsep}{4.5pt}
        \begin{tabular}{l cc cc cc cc cc}
        \toprule
        \multirow{2}{*}{\textbf{Method}} & \multicolumn{2}{c}{\textbf{Encoder}} & \multicolumn{2}{c}{\textbf{I2T}} & \multicolumn{2}{c}{\textbf{T2I}} & \multicolumn{2}{c}{\textbf{I2P}} & \multicolumn{2}{c}{\textbf{P2I}} \\
        \cmidrule(lr){2-3} \cmidrule(lr){4-5} \cmidrule(lr){6-7} \cmidrule(lr){8-9} \cmidrule(lr){10-11}
        & \textbf{Vision} & \textbf{Text} & \textbf{R@10} & \textbf{R@50} & \textbf{R@10} & \textbf{R@50} & \textbf{R@10} & \textbf{R@50} & \textbf{R@10} & \textbf{R@50} \\
        \midrule
        \rowcolor{gray!15}
        \multicolumn{11}{c}{\textbf{\textit{General VLMs}}} \\
        OpenCLIP \cite{openclip} & ViT-B & GPT2   & 10.72 & 24.25 & 10.08 & 22.95 & 2.88 & 8.34 & 2.79 & 8.24 \\
        SigLIP2 \cite{siglip2} & So400m & SigLIP64 & 14.02 & 28.16 & 10.33 & 23.44 & 0.31 & 0.76 & 0.08 & 0.62 \\
        CoCa \cite{coca} & ViT-B & GPT2     & 9.27  & 22.16 & 7.57  & 18.93 & 2.02 & 6.42 & 1.82 & 5.59 \\
        \midrule
        \rowcolor{gray!15}
        \multicolumn{11}{c}{\textbf{\textit{Biomedical VLMs}}} \\
        PMC-CLIP \cite{pmc-clip} & ViT-L & PubmedBert & 40.00 & 64.82 & 36.68 & 61.83 & 7.22 & 23.44 & 6.64 & 18.92 \\
        BiomedCLIP \cite{biomedclip} & ViT-B & PubmedBert & 32.91 & 56.63 & 32.43 & 56.08 & 3.71 & 13.38 & 3.77 & 12.17 \\
        BIOMEDICA \cite{biomedica} & ViT-L & GPT2  & 40.51 & 66.28 & 40.03 & 67.38 & 8.12 & 25.27 & 6.82 & 19.60 \\
        \midrule
        \rowcolor{green!15}
        \textbf{PhenoLIP} & ViT-B & PubmedBert & \textbf{63.30} & \textbf{81.92} & \textbf{66.61} & \textbf{87.68} & \textbf{13.84} & \textbf{36.88} & \textbf{12.77} & \textbf{31.30} \\
        \bottomrule
        \end{tabular}
        \vspace{-2mm}
        \caption{Cross-modal retrieval results on PhenoBench(7,819 images, 1,187 phenotypes). I2T represents image-to-text retrieval and T2I represents text-to-image retrieval. I2P represents image-to-phenotype retrieval and P2I represents phenotype-to-image retrieval. The best-performing model for each setting is in \textbf{bold}, the second-best is \underline{underlined}.}
        \label{tab:retrieval}
        \vspace{-4mm}
    \end{table*}

    \section{Results}
    In this section, we analyze the main results, comparing PhenoLIP with baselines across all evaluated settings, followed by ablation studies, as presented in Table~\ref{tab:zs}, \ref{tab:retrieval}, \ref{tab:face2gene}, and \ref{tab:lp}.

    \subsection{Zero-Shot Classification}
    We first benchmark models on their zero-shot accuracy in classifying medical images into task-related classes. 

    \vspace{2pt} \noindent \textbf{PhenoLIP unlocks superior zero-shot capabilities.}
    As shown in Table~\ref{tab:zs}, our {PhenoLIP} model substantially outperforms both general-domain and biomedical VLMs across a wide range of medical imaging modalities, demonstrating its superior capabilities in medical image understanding and perception. Notably, it achieves the highest average accuracy of 36.56\%, substantially surpassing the next best-performing model, {BiomedCLIP}~\cite{biomedclip}, which attained an accuracy of 27.41\%. The performance gain can be observed on 6 imaging modalities, indicating the model's strong generalization ability across diverse medical imaging domains.

    \vspace{2pt} \noindent \textbf{PhenoKG delivers robust cross-specialty generalization.}
    The strength of our approach is particularly evident in phenotype-rich domains. {In Radiology, PhenoLIP attains 49.03\% accuracy, creating a remarkable 11.31\% gap with the second-best model.} Strong performance is also seen in Pathology, where its 58.33\% accuracy represents a 0.93\% absolute improvement over BIOMEDICA. This trend of superior performance extends to other specialized fields like Hematology (23.24\%) and Ophthalmology (51.63\%). Conversely, the lack of improvement on the Histology (TissueMNIST) dataset is attributable to the nature of the task, which involves classifying broad tissue types rather than phenotype-wise abnormalities.

    \subsection{Cross-modal Retrieval.}
    In this section, we evaluate the cross-modal retrieval performance of PhenoLIP on two fronts: first, on the diverse PhenoBench dataset for both standard image-caption and phenotype-centric retrieval, and second, on the challenging Face2Gene~\cite{face2gene} benchmark for identifying rare phenotypes.

    \vspace{2pt}\noindent \textbf{PhenoLIP enhances cross‑modal retrieval performance on PhenoBench.}
    In Table~\ref{tab:retrieval}, we present the cross-modal retrieval results on PhenoBench, covering diverse medical image-caption pairs. Its diversity is summarized in Table \ref{tab:retrieval}.
    As shown by the results, in the standard I2T retrieval, PhenoLIP achieves an R@10 of 63.30\%, surpassing the second-best model, BIOMEDICA~\cite{biomedica}, by a remarkable margin of 22.79\%. Similar improvement can be observed in the T2I task, where our model's R@10 of 66.61\% exceeds BIOMEDICA's 40.03\%. This demonstrates that PhenoLIP has learned a significantly more accurate and robust alignment between various medical images and their corresponding textual descriptions.

    \vspace{2pt}\noindent \textbf{PhenoLIP enables better alignment between visual features and phenotypes.}
    Crucially, {PhenoLIP}'s advantage becomes even more pronounced in the more challenging and clinically meaningful phenotype-centric retrieval tasks. In the I2P retrieval setting, our model attains an R@10 of 13.84\%, nearly doubling the performance of the runner-up, {BIOMEDICA} (8.12\%). Similarly, for P2I retrieval, {PhenoLIP} reaches an R@10 of 12.77\%, markedly surpassing {BIOMEDICA}'s 6.82\%. The model's consistent and substantial lead in both I2P and P2I tasks, across a challenging vocabulary of 1187 distinct phenotype classes, confirms that our PhenoLIP method has successfully established a robust, bidirectional alignment between visual features and their specific clinical phenotype representations.

    \begin{table*}[!t]
        \footnotesize
        \centering
        \begin{adjustbox}{width=\textwidth}
        \setlength{\tabcolsep}{2.5pt}
        \begin{tabular}{lcccccccccccccccccccccccc}
            \toprule
        \multirow{2}{*}{\textbf{Method}} 
        & \multicolumn{3}{c}{\textbf{RSNA}} 
        & \multicolumn{3}{c}{\textbf{BreastMNIST}} 
        & \multicolumn{3}{c}{\textbf{ChestMNIST}} 
        & \multicolumn{3}{c}{\textbf{DermaMNIST}} 
        & \multicolumn{3}{c}{\textbf{OCTMNIST}} 
        & \multicolumn{3}{c}{\textbf{RetinaMNIST}} 
        & \multicolumn{3}{c}{\textbf{HAM10000}} \\
        \cmidrule(lr){2-4} 
        \cmidrule(lr){5-7} 
        \cmidrule(lr){8-10} 
        \cmidrule(lr){11-13} 
        \cmidrule(lr){14-16} 
        \cmidrule(lr){17-19} 
        \cmidrule(lr){20-22} 
            & \textbf{1\%} & \textbf{10\%} & \textbf{100\%} 
            & \textbf{1\%} & \textbf{10\%} & \textbf{100\%} 
            & \textbf{1\%} & \textbf{10\%} & \textbf{100\%} 
            & \textbf{1\%} & \textbf{10\%} & \textbf{100\%} 
            & \textbf{1\%} & \textbf{10\%} & \textbf{100\%} 
            & \textbf{1\%} & \textbf{10\%} & \textbf{100\%} 
            & \textbf{1\%} & \textbf{10\%} & \textbf{100\%} \\
        \midrule

        \multicolumn{22}{c}{\cellcolor{gray!15}\textbf{\textit{General VLMs}}} \\
        \midrule
        OpenCLIP \cite{openclip} 
            & 77.07 & 79.52 & 81.61
            & 32.05 & 73.72 & 83.33 
            & 51.31 & 52.88 & 53.69 
            & 69.53 & 77.51 & \underline{83.74} 
            & 67.40 & 73.20 & 73.10 
            & 43.50 & 55.00 & \underline{61.75} 
            & 52.81 & \underline{68.65} & \underline{77.23} \\
        SigLIP2 \cite{siglip2} 
            & 54.55 & 68.90 & 77.50
            & 32.50 & 73.50 & 83.00
            & 51.05 & 52.50 & 53.80
            & 69.00 & 77.10 & 83.50
            & 67.00 & 72.90 & 72.50
            & 43.80 & 54.50 & 61.00
            & 53.24 & 66.67 & 72.57 \\
        CoCa \cite{coca} 
            & 77.29 & 79.59 & 81.36
            & 36.54 & 78.21 & 82.05 
            & 51.62 & 52.85 & 53.48 
            & 68.73 & 75.26 & 81.40 
            & 43.75 & 55.75 & 59.50 
            & \textbf{52.24} & \underline{56.92} & 58.37
            & 55.45 & 66.67 & 72.94 \\
        \midrule

        \multicolumn{22}{c}{\cellcolor{gray!15}\textbf{\textit{Biomedical VLMs}}} \\
        \midrule
        PMC-CLIP \cite{pmc-clip} 
            & \underline{82.49} & \textbf{83.74} & 83.76
            & 26.92 & 75.00 & 82.05 
            & 49.50 & 53.70 & \textbf{55.19} 
            & 69.23 & 75.26 & 80.55 
            & \textbf{76.20} & 69.90 & 71.60 
            & \underline{45.00} & 56.50 & 60.00 
            & 54.13 & 64.36 & 75.25 \\
        BiomedCLIP \cite{biomedclip} 
            & 81.99 & \underline{83.66} & \underline{83.99}
            & \textbf{51.92} & \textbf{79.49} & 84.62 
            & 51.31 & \underline{54.46} & 55.03 
            & 69.28 & 74.86 & 79.25 
            & 68.20 & 73.80 & 74.10 
            & 43.50 & 55.75 & 59.00 
            & \underline{59.74} & 65.02 & 72.61 \\
        BIOMEDICA \cite{biomedica} 
            & 80.09 & 82.94 & 83.44
            & 37.82 & \underline{77.56} & \underline{85.90} 
            & \underline{51.99} & 53.57 & 54.45 
            & \textbf{70.87} & \textbf{79.10} & \textbf{85.14} 
            & 61.40 & \underline{76.40} & \underline{76.70} 
            & 43.50 & 55.50 & \textbf{63.50} 
            & 57.76 & \textbf{69.31} & 73.60 \\
        \midrule
        \rowcolor{green!15}
        \textbf{PhenoLIP} 
            & \textbf{83.06} & 83.36 & \textbf{84.11}
            & \underline{49.36} & 74.36 & \textbf{86.54} 
            & \textbf{52.65} & \textbf{54.71} & \underline{55.15} 
            & \underline{69.93} & \underline{75.86} & \underline{83.74} 
            & \underline{73.90} & \textbf{78.30} & \textbf{77.10} 
            & 43.50 & \textbf{57.75} & \underline{61.75} 
            & \textbf{62.71} & \textbf{69.31} & \textbf{77.89} \\
        \bottomrule
        \end{tabular}
        \vspace{-12mm}
        \end{adjustbox}
        \caption{Benchmarking results on Linear Evaluation (Acc) across different data ratios. The best-performing model for each setting is in \textbf{bold}, and the second-best is \underline{underlined}.}
        \label{tab:lp}
        \vspace{-2mm}
    \end{table*}
    % face
    \begin{table}[t]
        \footnotesize
        \centering
        \setlength{\tabcolsep}{4pt} % 可以根据页面宽度微调列间距
        \renewcommand{\arraystretch}{1.1} % 调整行间距
        % 列定义减少到7列 (l + 6c)
        \begin{tabular}{lcccccc}
        \toprule
        % --- 表头现在只有7列 ---
        \multirow{2}{*}{\textbf{Method}} & \multicolumn{3}{c}{\textbf{Retrieval Metrics (\%)}} & \multicolumn{3}{c}{\textbf{Matching Metrics (\%)}} \\
        % cmidrule的列索引和跨度已更新
        \cmidrule(lr){2-4} \cmidrule(lr){5-7}
        % 移除了 R@20
        & \textbf{R@5} & \textbf{R@10} & \textbf{R@50} & \textbf{Prec.} & \textbf{Recall} & \textbf{F1} \\
        \midrule
        % --- multicolumn 的跨度减少到7列 ---
        \multicolumn{7}{c}{\cellcolor{gray!15}\textbf{\textit{General VLMs}}} \\
        \midrule
        % --- 数据行中移除了 R@20 的值 ---
        OpenCLIP \cite{openclip} & \underline{6.97} & \underline{11.94} & 39.80 & \underline{1.63} & 2.56 & \underline{1.81} \\
        SigLIP2 \cite{siglip2} & 6.85 & 11.80 & 39.50 & 1.50 & 2.10 & 1.75 \\
        CoCa \cite{coca} & 4.48 & 7.46 & 38.31 & 1.04 & 2.21 & 1.25 \\
        \midrule
        % --- 第二个分组标题 ---
        \multicolumn{7}{c}{\cellcolor{gray!15}\textbf{\textit{Biomedical VLMs}}} \\
        \midrule
        PMC-CLIP \cite{pmc-clip} & 6.75 & 11.60 & 49.50 & 1.65 & 3.20 & 2.17 \\
        % 注意：BiomedCLIP 的 R@20 是加粗的最佳值，移除后不再显示
        BiomedCLIP \cite{biomedclip} & 6.84 & 11.73 & \underline{49.84} & \underline{1.80} & \underline{3.77} & \underline{2.22} \\
        BIOMEDICA \cite{biomedica} & 3.48 & 7.46 & 43.28 & 1.10 & 2.37 & 1.36 \\
        \midrule
        % 注意：PhenoLIP 的 R@20 是下划线的次佳值，移除后不再显示
        \rowcolor{green!15}
        \textbf{PhenoLIP} & \textbf{7.49} & \textbf{12.05} & \textbf{55.05} & \textbf{2.08} & \textbf{4.56} & \textbf{2.62} \\
        \bottomrule
        \end{tabular}
        \vspace{-2mm}
        \caption{Rare facial phenotype identification results on the Face2Gene~\cite{face2gene} dataset (321 images, average phenotypes per image:21, 993 candidate phenotypes). The best result per column is in \textbf{bold}, and the second-best is \underline{underlined}.}
        \vspace{-6mm}
        \label{tab:face2gene}
    \end{table}
    % \vspace{-3pt}

    \vspace{1pt} \noindent \textbf{PhenoLIP demonstrates better perception of rare phenotypes.}
    To assess the model's ability to identify rare phenotypes from the long tail of the distribution, we introduce a rare phenotype retrieval task, linking medical facial images with related phenotype terms. As shown in Table \ref{tab:face2gene}, PhenoLIP consistently outperforms the other evaluated methods across all metrics in this highly challenging setting. The model's advantage is most evident in the retrieval, where it achieves the highest scores with an R@5 of 7.49\% and R@50 of 55.05\%. It also leads in direct phenotype matching, attaining the best F1-score of 2.62\%. The low absolute matching scores across all models reflect the task's extreme difficulty, as each image has, on average, 21 related phenotypes within a vast pool of 993 facial phenotype candidates. Despite this challenge, PhenoLIP's consistent top-ranking performance demonstrates its enhanced ability to capture subtle visual signatures of rare diseases.

    \subsection{Linear Probing}
    Table~\ref{tab:lp} presents the linear probing results, which evaluate the quality and transferability of the learned visual representations. Our PhenoLIP model demonstrates remarkable performance across 21 evaluation settings (7 datasets at three data ratios), PhenoLIP achieves either the best or second-best performance in 19 instances.

    The representations from PhenoLIP demonstrate exceptional data efficiency in low-data scenarios. For instance, using just 1\% of the labels for the ChestMNIST and HAM10000 datasets, it achieved the top position with accuracies of 52.65\% and 62.71\%, respectively. As the data ratio increases to 10\%, its superiority in the ophthalmology modality becomes prominent, achieving first place on both OCTMNIST and RetinaMNIST with 78.30\% and 57.75\% accuracy. When using the full 100\% dataset, the model sustains its leading performance, achieving the top rank on several datasets, including RSNA, BreastMNIST, and HAM10000, with respective accuracies of 84.11\%, 86.54\%, and 77.89\%. These leading results across diverse modalities, spanning X-ray, dermatoscopy, ophthalmology, and ultrasound, provide compelling evidence that the representations learned by PhenoLIP are highly effective and robust in low-data regimes and high-data scenarios, indicating excellent generalization capabilities.

    % ablation
    \begin{table}[!t]
        \footnotesize
        \centering
        \setlength{\tabcolsep}{3pt}
        \begin{tabular}{cccccccccc}
        \toprule
        \multicolumn{2}{c}{\textbf{Encoder}} & \textbf{Know.} & \textbf{Data} & \multicolumn{2}{c}{\textbf{I2T}} & \multicolumn{2}{c}{\textbf{T2I}} & \textbf{Cls.} \\
        \cmidrule(lr){1-2} \cmidrule(lr){3-3} \cmidrule(lr){4-4} \cmidrule(lr){5-6} \cmidrule(lr){7-8} \cmidrule(lr){9-9}
        \textbf{Vision} & \textbf{Text} & \textbf{Distill.} & \textbf{Cur.} & \textbf{R@10} & \textbf{R@50} & \textbf{R@10} & \textbf{R@50} & \textbf{Acc} \\ 
        \midrule
        Scratch & KB & & & 28.15 & 45.33 & 31.04 & 49.81 & 2.13 \\
        CLIP & KB & & & 47.20 & 68.91 & 51.72 & 73.05 & 6.44 \\ 
        BCLIP & KB & & & 50.11 & 71.22 & 53.68 & 75.90 & 7.02 \\ 
        CLIP & PMB & & & 49.53 & 70.88 & 54.88 & 78.14 & 6.95 \\
        BCLIP & PMB & & & 53.42 & 74.10 & 58.03 & 81.25 & 7.81 \\ 
        CLIP & PMB & \checkmark & & 58.91 & 78.54 & 62.19 & 84.33 & 9.57 \\
        CLIP & PMB & \checkmark & \checkmark & 60.13 & 79.62 & 63.55 & 85.18 & 9.98 \\
        \rowcolor{green!15}
        BCLIP & PMB & \checkmark & \checkmark & \textbf{63.30} & \textbf{81.92} & \textbf{66.61} & \textbf{87.68} & \textbf{10.76} \\
        \bottomrule
        \end{tabular}
        \vspace{-2mm}
        \caption{Ablation study results. We evaluate the impact of {different encoder initializations}, {knowledge distillation}, and {data curation strategies} on image-to-text (I2T) retrieval, text-to-image (T2I) retrieval, and classification performance. BCILP denotes vision encoder of BiomedCLIP, PMB and KB denotes PubmedBert and our pretrained knowledge encoder. Our full model, combining all components, achieves the best results across all tasks. The Best results and related configurations are in \textbf{bold}.}
        \label{tab:ablation}
    \end{table}

    \subsection{Ablation Studies}
    To demonstrate the impact of each component of our method, we conduct a series of ablation studies with the cross-modal retrieval task on PhenoBench for: (\textit{i}) different encoder initializations, (\textit{ii}) our proposed knowledge distillation loss, and (\textit{iii}) data curation strategies. As shown in Table~\ref{supptab:ablation}, we can draw the following key observations (detailed discussion is listed in Appendix~\ref{suppsec:ablation}):

    \begin{itemize}
    \setlength\itemsep{3pt}
        \item \textbf{Encoder initialization:} Domain-specific pre-trained encoders, such as BiomedCLIP and PubmedBERT, serve as stronger baselines, confirming the value of biomedical-informed representations.
        \item \textbf{Knowledge distillation loss:} The proposed $\mathcal{L}_{\text{KD}}$ contributes significantly to performance gains. By aligning the VLM's text encoder with the frozen phenotype knowledge encoder, it effectively injects structured ontological knowledge into the model.
        \item \textbf{Data curation strategies:} Our fine-grained data curation pipeline, including subfigure detection and LLM-based caption refinement, consistently improves model performance. These steps reduce noise from compound figures and enhance alignment between images and text, leading to more robust vision-language representations.
        {
        \item \textbf{Knowledge graph components:} We further evaluate the contribution of different components (definitions, synonyms, and relational structure) in the phenotype knowledge encoder, detailed in Appendix~\ref{suppsec:kg_ablation}. The results show that both semantic definitions and hierarchical graph topology significantly contribute to the overall vision-language pretraining.
        }
    \end{itemize}

%% file: sec/5_conclusion.tex
\section{Conclusion}
\label{sec:conclusion}
In this work, we presented \textbf{PhenoLIP}, a phenotype-ontology-aware vision–language pretraining framework that integrates structured phenotype knowledge into medical VLMs. Built upon our newly curated large-scale multimodal knowledge graph \textbf{PhenoKG} and evaluated with the expert-verified benchmark \textbf{PhenoBench}, PhenoLIP achieves substantial gains in phenotype recognition, diagnostic classification, and retrieval tasks. These results highlight the importance of combining medical VLMs with phenotype‑centric knowledge to achieve more accurate and interpretable medical image understanding.

%% file: sec/X_suppl.tex
\onecolumn
\clearpage
\tableofcontents
\clearpage
\setcounter{section}{7}
\section{Qualitative Analysis}
                        
In this section, we provide qualitative analyses on PhenoBench. By visually comparing the top-10 retrieval results of PhenoLIP and BiomedCLIP for the same phenotype queries, we show that PhenoLIP retrieves more correct and phenotypically relevant images, highlighting its advantage in fine-grained phenotype understanding.

\begin{figure}[h]
    \centering
    \includegraphics[width=1\linewidth]{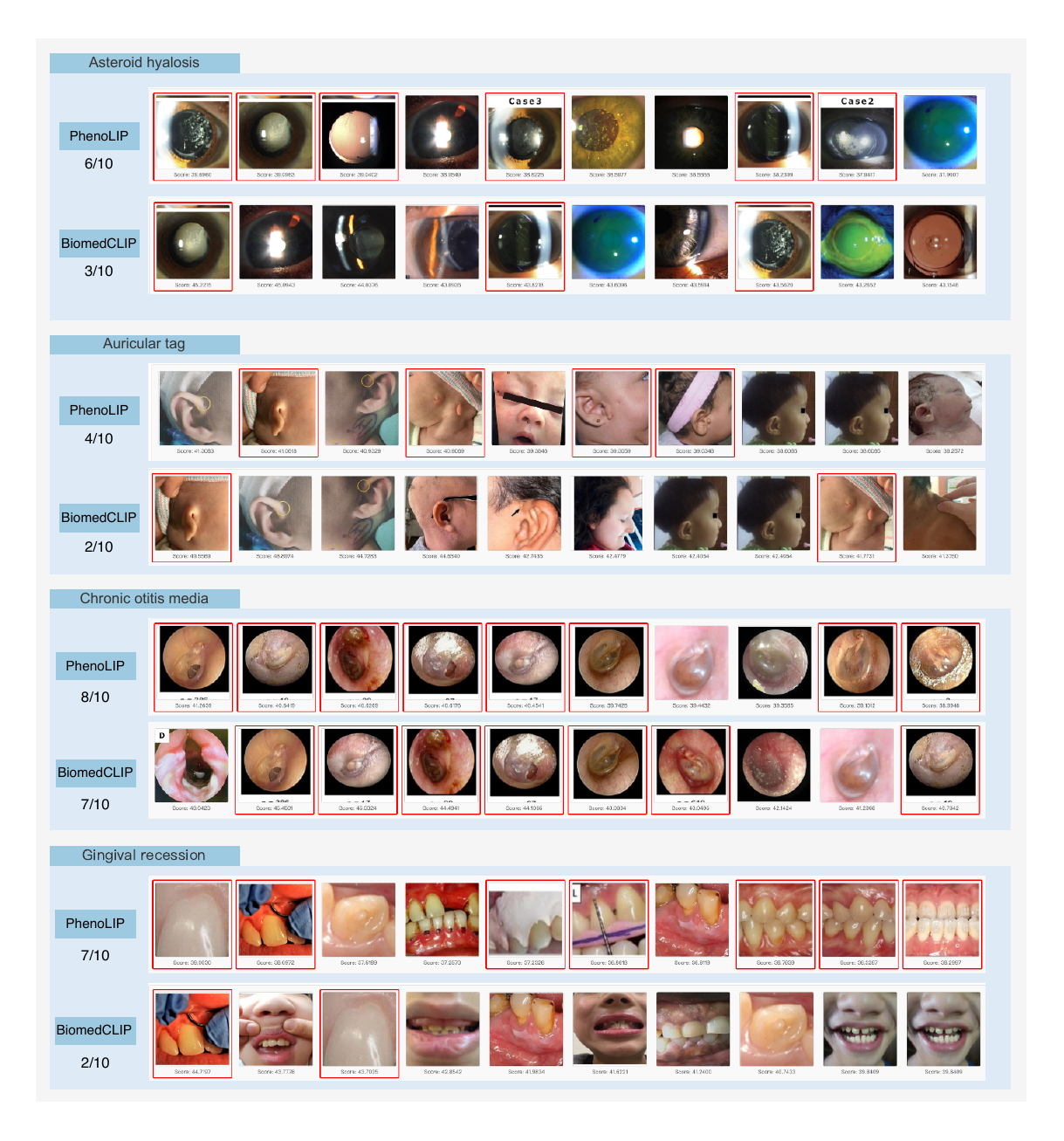}
    \caption{Comparison of PhenoLIP and BiomedCLIP on phenotype retrieval from PhenoBench. For each phenotype query (left), we show the top-10 retrieved images from both models. Correct retrievals are highlighted with red boxes.}
    \label{suppfig:retrieval}
\end{figure}

\vspace{3pt} \noindent \textbf{Cross-modal retrieval examples.} In Figure~\ref{suppfig:retrieval}, we show the qualitative examples of cross-modal retrieval on PhenoBench, comparing PhenoLIP with BiomedCLIP. As shown by the results, our proposed PhenoLIP can retrieve more accurate and relevant images in response to the phenotype queries, demonstrating its superior understanding of fine-grained visual features.

\section{PhenoKG Statistics}
\label{sec:stats}
\subsection{Ontology Statistics}
Table~\ref{tab:ontology_stats} demonstrates the statistics of the ontology terms considered in PhenoKG. Our knowledge graph comprises \textbf{19,703 phenotype nodes}, which are interconnected through \textbf{23,528 \texttt{is\_a} relationships}. 
Furthermore, we incorporate 14,020 synonyms covering approximately 50\% of the specific phenotypes. In total, the ontology provides 27,832 unique textual phenotype descriptions, forming a comprehensive foundation for phenotype knowledge injection.

\begin{table}[h]
    \centering
    \begin{tabular}{lr}
    \toprule
    \textbf{Item} & \textbf{Count} \\
    \midrule
    Number of phenotypes & 19,703 \\
    Number of \texttt{is\_a} relationships & 23,528 \\
    Number of phenotype groups & 5,839 \\
    Number of specific phenotypes & 13,812 \\
    Number of specific phenotypes with synonyms & 6,990 \\
    Number of synonyms for specific phenotypes & 14,020 \\
    Number of entries (phenotypes + synonyms) & 27,832 \\
    \bottomrule
    \end{tabular}
    \caption{Detailed statistics of the textual PhenoKG ontology. The ontology distinguishes between broad phenotype groups and specific phenotypes, enriched with hierarchical relations and synonyms to support robust knowledge encoding.}
    \label{tab:ontology_stats}
\end{table}

\subsection{PhenoKG Multimodal Statistics}
\label{suppsec:phenokg_stat}
\begin{figure}[h]
    \centering
    \begin{minipage}[t]{0.48\textwidth}
        \centering
        \includegraphics[width=\linewidth]{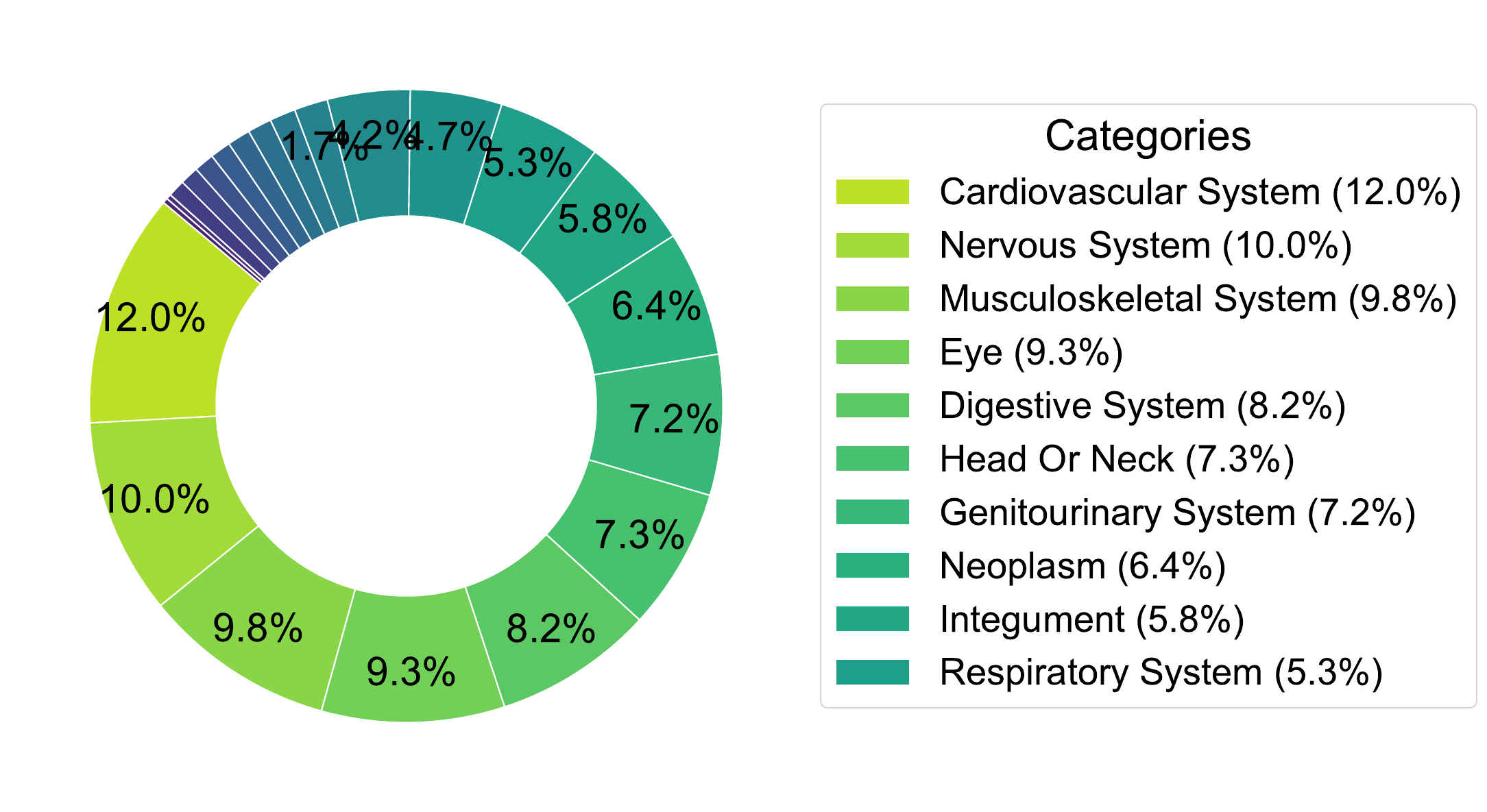}
        \label{fig:phenokg_category}
    \end{minipage}
    \hfill
    \begin{minipage}[t]{0.48\textwidth}
        \centering
        \includegraphics[width=\linewidth]{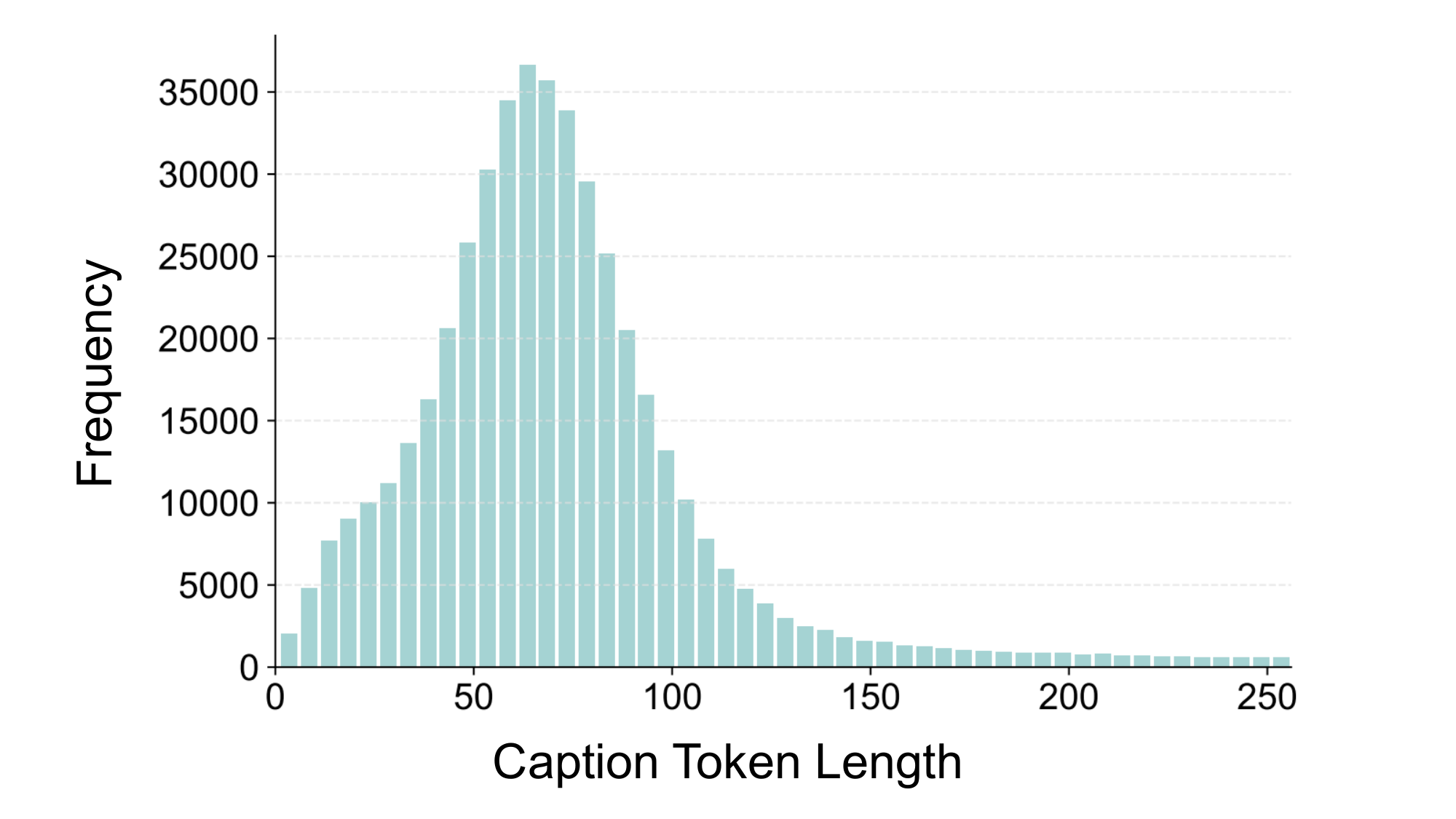}
        \label{fig:phenokg_token}
    \end{minipage}
    \caption{(left) Distribution of phenotype categories in PhenoKG. The dataset spans a broad range of anatomical systems, with cardiovascular, nervous, and musculoskeletal phenotypes among the most frequently represented. (right) Distribution of caption lengths (in tokens) in PhenoKG. Most captions are comprehensive, with a mean of 63 tokens, indicating that the image–text pairs provide rich and detailed clinical descriptions.}
    \label{fig:phenokg_category_token}
\end{figure}

Our final \textbf{PhenoKG} consists of 3,096 unique multimodal phenotype nodes, augmented with 524,804 aligned image–caption pairs. As shown in Figure~\ref{fig:phenokg_category_token}(left), these phenotypes span a wide range of anatomical systems.
Among these systems, cardiovascular, nervous, and musculoskeletal abnormalities contribute a substantial portion, while the remaining systems also well-represented.
In Figure~\ref{fig:phenokg_category_token}(right), we further show the caption token length distribution, with a mean of 63 tokens, indicating that most images are accompanied by detailed clinical narratives rather than short keywords or brief labels.

\begin{figure}[h]
    \centering
    \begin{minipage}[t]{0.48\textwidth}
        \centering
        \includegraphics[width=\linewidth]{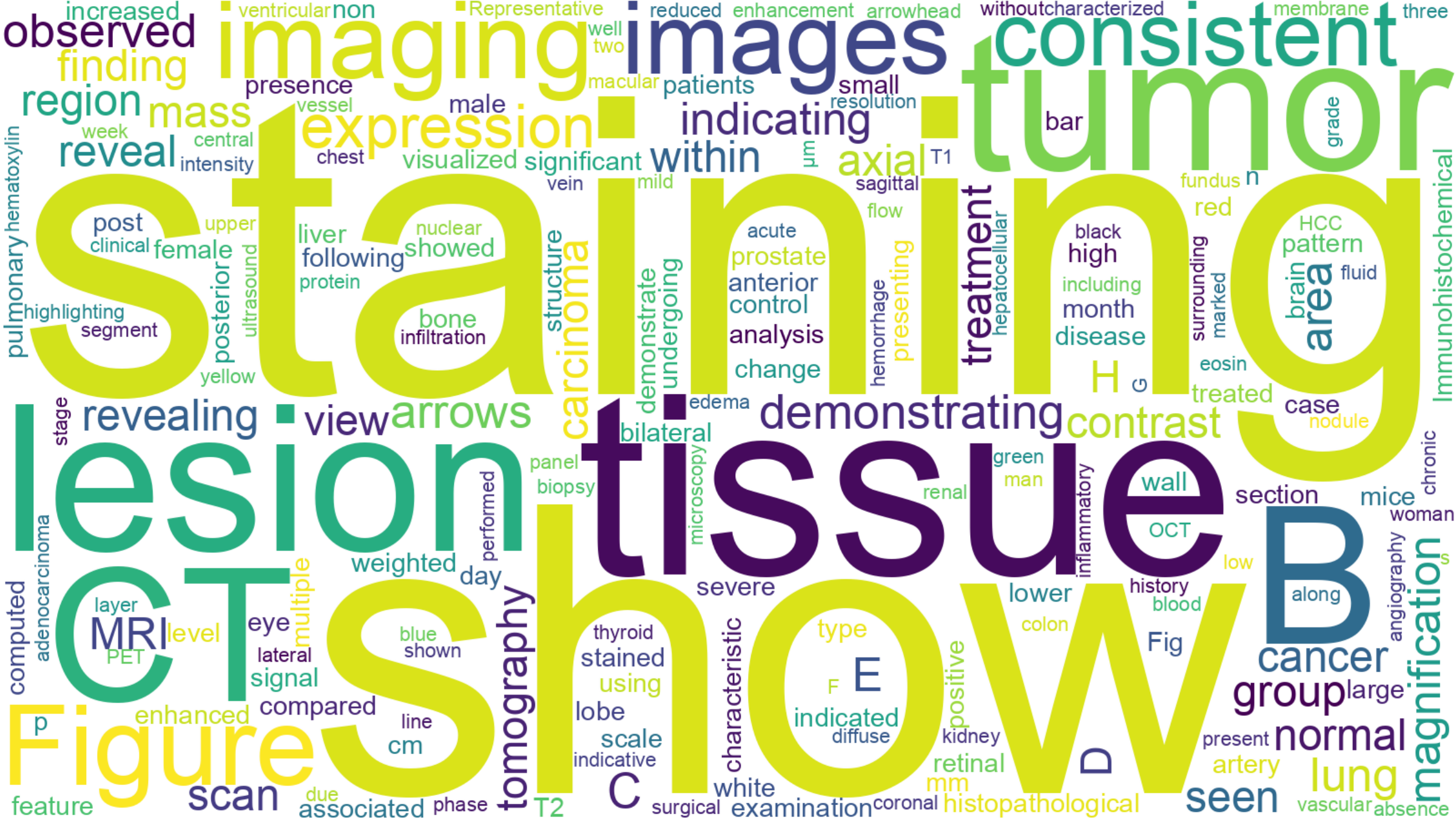}
        \label{fig:phenokg_wordcloud}
    \end{minipage}
    \hfill
    \begin{minipage}[t]{0.48\textwidth}
        \centering
        \includegraphics[width=\linewidth]{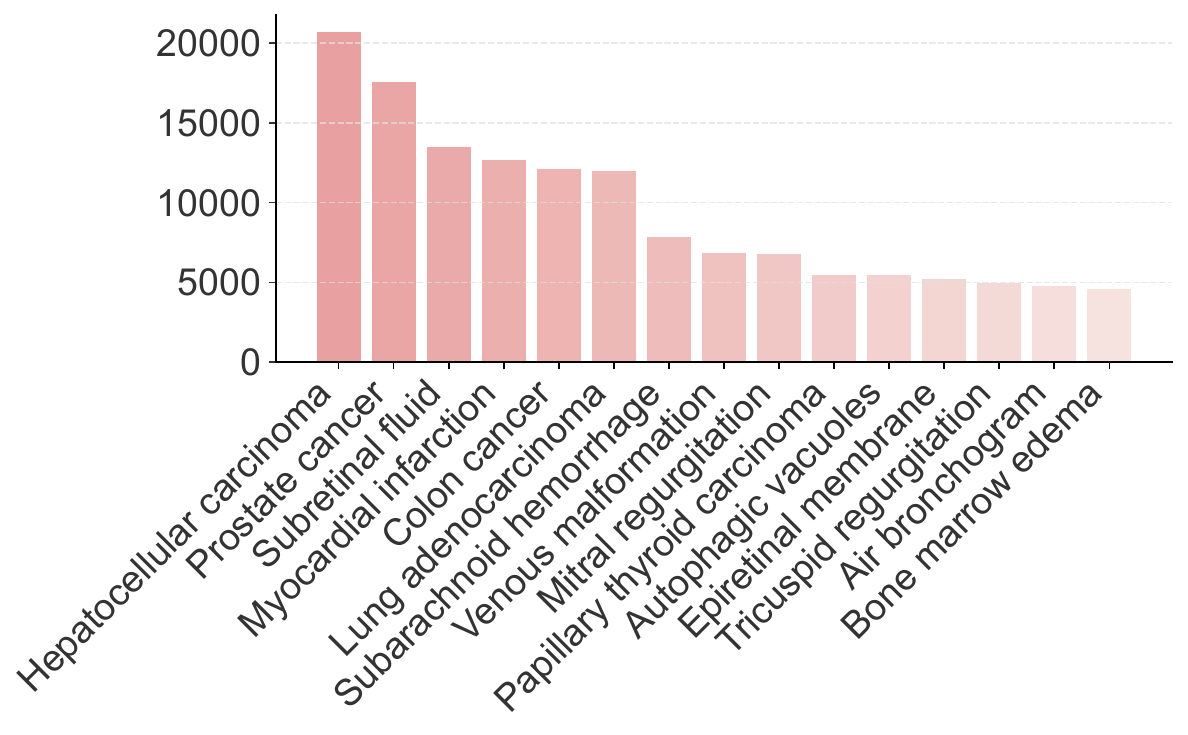}
        \label{fig:phenokg_top15}
    \end{minipage}
    \caption{(left) Word cloud of the caption corpus in PhenoKG. Frequent terms include phenotype names, anatomical structures, and clinical descriptors, reflecting the diverse and fine-grained medical semantics captured in the dataset. (right) Frequency distribution of the top-15 most common phenotypes in PhenoKG. }
    \label{fig:phenokg_word_top15}
\end{figure}

Moreover, as shown by the word cloud in Figure~\ref{fig:phenokg_word_top15}(left), the caption vocabulary is dominated by visual descriptive expressions and medical terms.
The frequency distribution of the top-15 phenotypes is shown in Figure~\ref{fig:phenokg_word_top15}(right), which exhibits a clear long-tail pattern where a small set of common phenotypes appear frequently, while most phenotypes occur rarely. This pattern indicates that our PhenoKG includes both common and rare clinical presentations, which is important for training models that generalize to diverse and low-frequency medical image analysis conditions.

The overall statistics of PhenoKG are as follows:
\begin{table}[h]
    \centering
    \begin{tabular}{p{0.5\columnwidth} r}
    \toprule
    \textbf{Item} & \textbf{Count} \\
    \midrule
    Number of image-text pairs & 524,804 \\
    Number of unique phenotypes in PhenoKG & 19,703 \\
    Number of unique phenotypes paired with image-caption & 3,096 \\
    Average caption length (in tokens) & 63 \\
    \bottomrule
    \end{tabular}
    \caption{Overall statistics of PhenoKG.}
    \label{tab:phenokg_stats}
\end{table}

\section{Human-in-the-Loop Quality Control}
\label{sec:qc}
To ensure the quality of PhenoKG, we performed careful manual quality control throughout the data‑curation pipeline.

\subsection{Manual Evaluation of Image Filtering}
\label{sec:kmeans}

\begin{figure}[!h]
    \centering
    \includegraphics[width=0.9\linewidth]{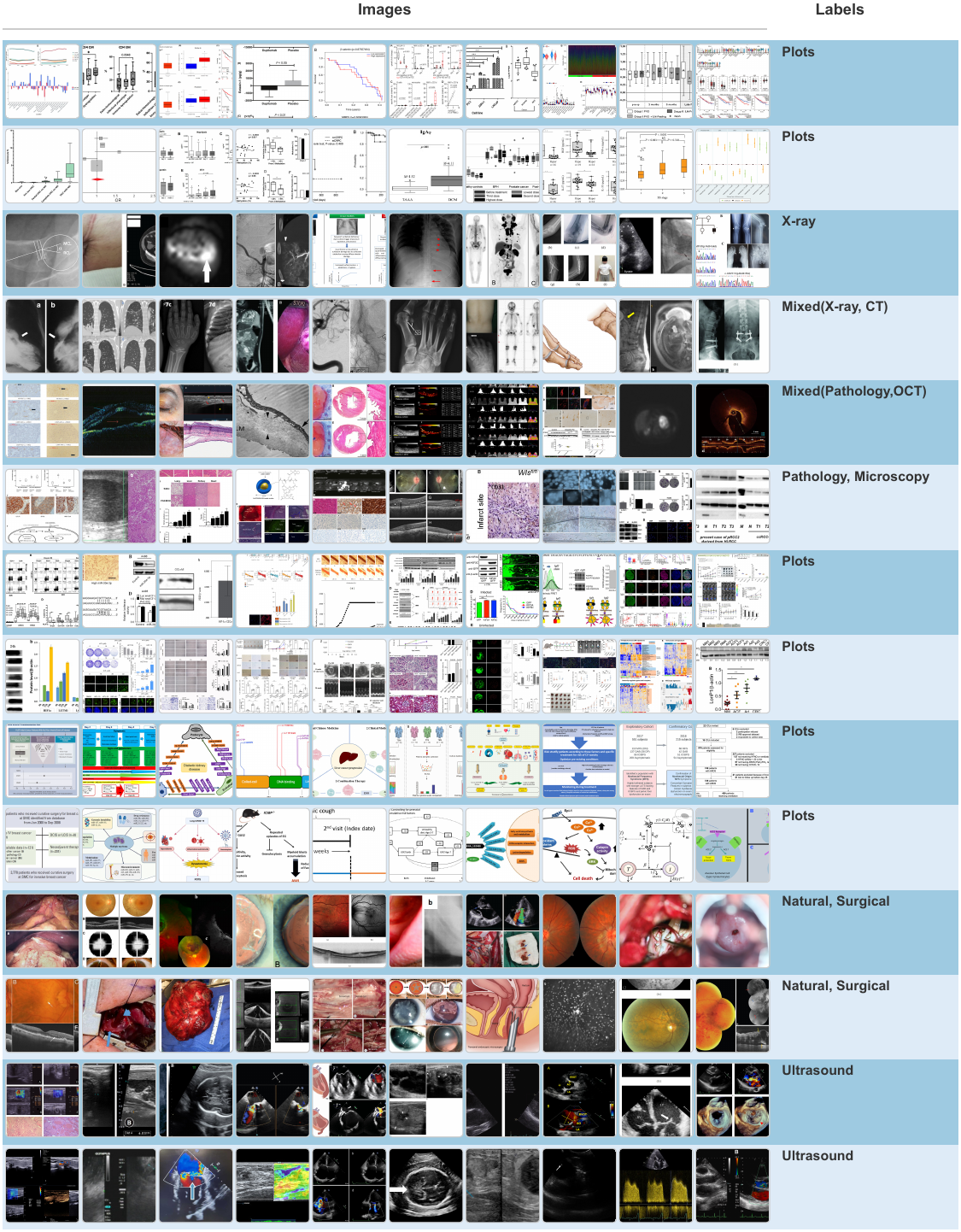}
    \caption{Visualization of K-means clusters used in our filtering pipeline. For each cluster, we sample 10 representative images and manually assign a category label. Clusters dominated by icons, charts, or other non-medical figures are removed, while clusters composed of medical images are retained.}
    \label{fig:placeholder}
\end{figure}

\begin{figure}[h]
    \centering
    \includegraphics[width=0.9\linewidth]{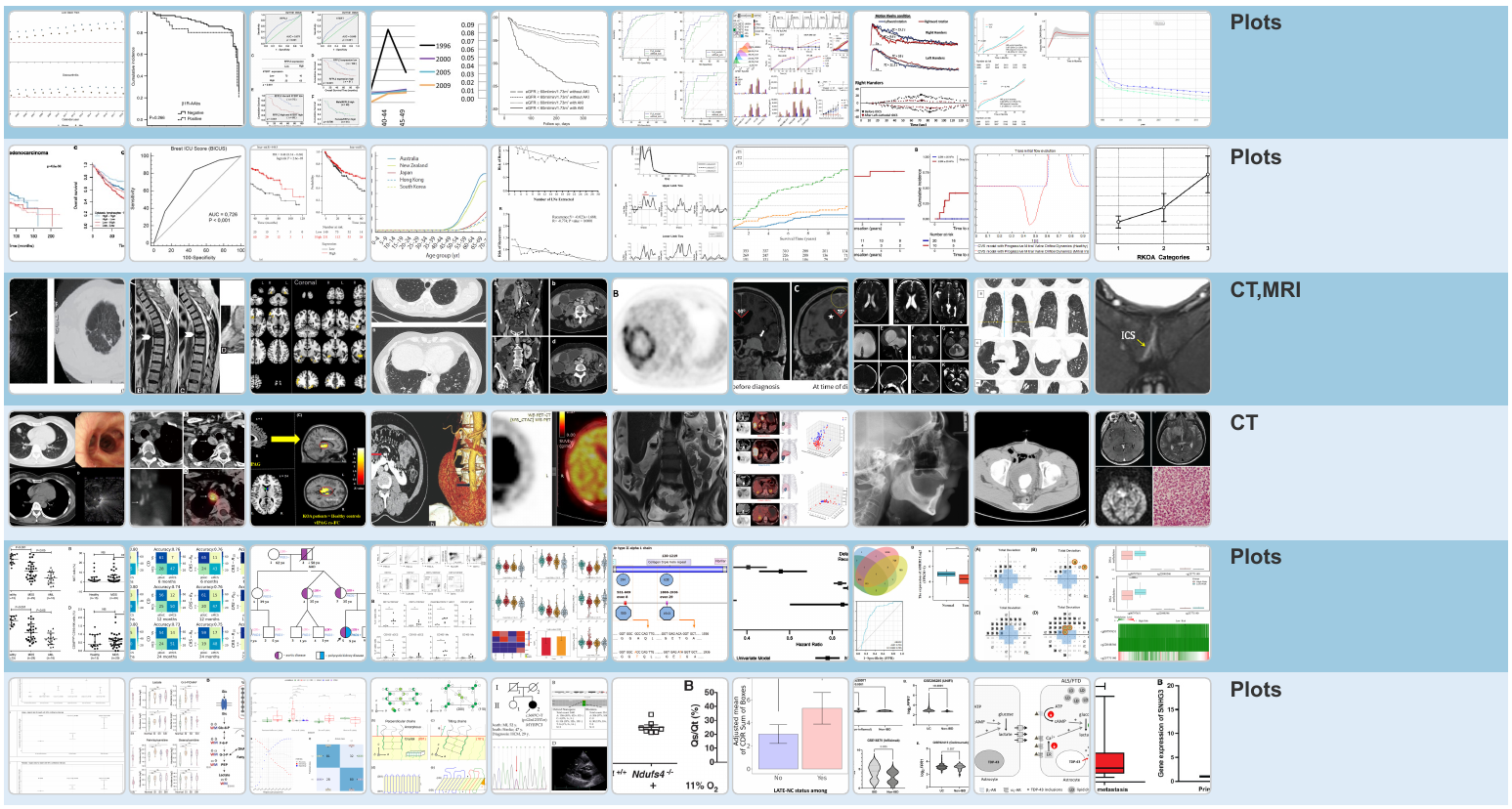}
    \caption{Continuation of the visualization of K-means clusters used in our filtering pipeline. }
    % \label{fig:placeholder}
\end{figure}

In the data filtering stage, we aim to remove non-medical figures (\emph{e.g.}, icons, charts, logos) and noisy images, retaining medical images. We first extract image features using DINOv3 and perform a two-level K-means clustering. At the first level, all images are clustered into $k=20$ groups. Each group is then further clustered into $k=20$ subgroups, resulting in a total of 400 clusters.

To clean the data, we manually audit these clusters. For each cluster, we randomly sample 10 representative images and manually assign a semantic label describing the predominant content (\emph{e.g.}, pathology, X-ray, icon, chart, \emph{etc.}). Clusters dominated by non-medical or irrelevant content, like icons, diagrams, charts, logos, \emph{etc.}, are discarded. Figure~\ref{fig:placeholder} visualizes representative samples from selected clusters together with their manually assigned labels, illustrating that our clustering‑based pipeline can naturally distinguish medical images.

We then validate how well this pipeline retains medical images and filters out non-medical ones. We randomly sample 500 images from the original unfiltered data. Clinical experts manually annotate each image as either medically relevant or non-medical. We run the same filtering pipeline on these 500 images and compare its binary decisions (kept vs.\ removed) against the expert annotations, and compute precision and recall for identifying medical images. Our clustering-based filter achieves a precision of \textbf{96.2\%} and a recall of \textbf{93.5\%}, indicating that non-medical figures are reliably removed while medical ones are effectively preserved.

\subsection{Manual Evaluation of Subfigure Detection}
\label{sec:subfigure_detection}
For the subfigure detection step, we performed a manual evaluation of the DAB-DETR~\cite{dabdetr} model on a random sample of 200 compound figures from our raw image caption data. Each figure was independently annotated by clinical experts, who marked all visible subfigures with bounding boxes and associated panel identifiers (\emph{e.g.}, (a), (b), (c)). We then compared the bounding boxes predicted by the DAB-DETR model against the human annotation. Following standard object detection practice, we computed precision, recall, and F1-score at an IoU threshold of 0.5, treating each subfigure bounding box as a detection target. On this manually annotated set, DAB-DETR achieves a precision of \textbf{94.1\%}, a recall of \textbf{92.7\%}, and an F1-score of \textbf{93.4\%}, indicating that most subfigures are correctly localized with accurate extents.

\subsection{Manual Evaluation of Subfigure Alignment}
\label{sec:subfigure_align}
To validate the effectiveness of our subfigure–text realignment procedure, we conducted a manual review on a random sample of 300 subfigure–text pairs generated in our data curation procedure. Each realigned pair was evaluated by clinical experts to assess whether the textual description exactly corresponded to the visual content of the subfigure. Reviewers rated each pair as either ``accurate'' or ``inaccurate'' based on the degree of semantic alignment between the image and the text. Based on expert judgments, we computed accuracy, as well as precision, recall, and F1-score by treating expert labels as ground truth and our pipeline's assignments as predictions. Our pipeline achieves an accuracy of \textbf{92.7\%}, a precision of \textbf{93.4\%}, a recall of \textbf{92.1\%}, and an F1-score of \textbf{92.8\%}, indicating that the vast majority of subfigure–text pairs are judged to be correctly aligned. These findings demonstrate that, after subfigure detection and LLM-based text refinement, our subfigure–text alignment strategy yields high‑quality results for denoising.

\begin{table*}[h]
    \centering
    \caption{Evaluation benchmark datasets across different medical imaging domains. Each row lists the dataset name, task description, image modality, and number of classes.}
    \begin{adjustbox}{max width=\textwidth}
    \begin{tabular}{p{3cm}p{5cm}p{4.5cm}p{2cm}}
    \toprule
     \textbf{Dataset} & \textbf{Task} & \textbf{Modality} & N. Classes \\ \midrule
    PhenoBench & Phenotype retrieval / recognition & Various & 1187 \\
    Face2Gene~\cite{face2gene} & Facial phenotype retrieval & Face Photos & 299 \\

    \midrule
    \rowcolor{gray!10}\textbf{Dermatology} &&& \\
    DermaMNIST~\cite{medmnistv2} & Skin lesion classification & Dermatoscope & 7 \\
    HAM10000~\cite{ham10000} & Skin lesion diagnosis & Dermatoscope & 7 \\

    \midrule
    \rowcolor{gray!10} \textbf{Radiology} &&& \\ 
    ChestMNIST~\cite{medmnistv2} & Chest anomaly detection & Chest X-Ray & 14 \\
    BreastMNIST~\cite{medmnistv2} & Breast cancer diagnosis & Breast Ultrasound & 2 \\
    RSNA~\cite{cxr8} & Chest pathology classification & Chest X-Ray & 2 \\

    \midrule
    \rowcolor{gray!10}\textbf{Pathology} &&& \\
    LC25000~\cite{lc25000} & Lung and colon tissue classification & Light Microscopy (H\&E) & 5 \\

    \midrule
    \rowcolor{gray!10} \textbf{Ophthalmology} &&&\\
    OCTMNIST~\cite{medmnistv2} & Retinal disease detection & Optical Coherence Tomography & 4 \\
    RetinaMNIST~\cite{medmnistv2} & Diabetic retinopathy grading & Retina Fundus Images & 5 \\

    \midrule
    \rowcolor{gray!10}\textbf{Hematology} &&& \\
    BloodMNIST~\cite{medmnistv2} & Blood cell classification & Light Microscopy & 8 \\

    \midrule
    \rowcolor{gray!10}\textbf{Histology} &&& \\
    TissueMNIST~\cite{medmnistv2} & Tissue type classification & Light Microscopy & 8 \\

    \bottomrule
    \end{tabular}
    \end{adjustbox}
    \label{table:benchmark_sources}
\end{table*}
\section{Downstream Datasets}
\label{suppsec:downstream_datasets}

As shown in Table~\ref{table:benchmark_sources}, we selected eight publicly available medical image classification datasets from various medical imaging domains, covering dermatology, radiology, pathology, ophthalmology, hematology, and histology. Each dataset provides high-quality images and corresponding labels, making them suitable for evaluating model performance across diverse medical imaging tasks. The specific datasets include DermaMNIST, HAM10000, ChestMNIST, BreastMNIST, LC25000, OCTMNIST, RetinaMNIST, and BloodMNIST. Detailed information about these datasets, including task descriptions, image modalities, and the number of classes, is provided in Table~\ref{table:benchmark_sources}.

\section{Prompt Details in Data Curation}

In this section, we introduce the prompts used in our data curation pipeline to instruct LLMs or multimodal LLMs in detail.

\subsection{Text Processing Prompts}
\label{sec:text_splitting}
To obtain precise subfigure-level textual descriptions, we design a dedicated instruction prompt that guides LLMs to decompose the original figure caption into multiple sub-figure–specific captions. As shown in the box below, the model is explicitly instructed to (i) strictly separate content belonging to different subfigures, (ii) minimize shared contextual information, and (iii) emphasize accurate visual localization and clinical terminology. The LLM receives both the main caption and the reference paragraph as input, and is required to output a structured JSON object that maps each subfigure identifier to an enhanced caption and its corresponding imaging modality.

\begin{tcolorbox}[
    colback=gray!5,
    colframe=gray!60,
    title={Subfigure Alignment Prompt},
    breakable
]
\begin{lstlisting}[
    basicstyle=\small\ttfamily,
    columns=fullflexible,
    breaklines=true,
    breakatwhitespace=true
]
System:
You are a medical AI expert specializing in precise analysis of medical sub-figures. Your task is to create accurate, sub-figure-specific descriptions by carefully distinguishing the unique content of each individual sub-figure.

**You are given:**
1. **Main Caption:**
    ---
    {main\_caption}
    ---
2. **Reference Paragraph:**
    ---
    {reference\_para}
    ---

**Critical Requirements:**

**Principle 1: Strict Content Separation**
- Each sub-figure must ONLY describe visual elements present in THAT specific sub-figure
- NEVER attribute findings from one sub-figure to another
- If uncertain about which sub-figure contains a specific finding, describe it generally or omit it

**Principle 2: Minimal Shared Context**
- Start with ONLY essential background (1 sentence): patient condition, anatomical region, or study purpose
- Do NOT repeat detailed findings that apply to other sub-figures

**Principle 3: Precise Visual Localization**
- Use positional language: "upper left image shows...", "right panel demonstrates...", "lower image reveals..."
- Reference specific sub-figure identifiers when provided: "Sub-figure A displays...", "Panel B indicates..."

**Step-by-Step Process:**

**Step 1: Sub-figure Mapping**
- Identify each sub-figure using letters (A, B, C), numbers, or position (upper/lower/left/right)
- Note which specific visual content belongs to each sub-figure

**Step 2: Content Allocation**
- Carefully assign each visual finding to its correct sub-figure
- When in doubt, be conservative and describe only what's clearly visible

**Step 3: Focused Description Generation**
For each sub-figure, create descriptions that:
- Begin with minimal essential context only
- Focus exclusively on that sub-figure's unique visual content
- Use precise anatomical and imaging terminology
- Include specific imaging modality and view details
- **Length Constraint: Keep each enhanced\_caption under 256 tokens (approximately 150-200 words)**

**Output Format:**
Return a single, valid JSON object. Each key should be the sub-figure identifier ("A", "B", "main", etc.), and the value should be an object containing the `enhanced\_caption` and `modality`.

**Example Output Format:** :
```json
{
    "A": {
        "enhanced\_caption": "Focused description of sub-figure A content only...",
        "modality": "MRI"
    },
    "B": {
        "enhanced\_caption": "Specific description of sub-figure B content only...", 
        "modality": "CT"
    }
}
```
Generate the JSON output now: /no\_think
\end{lstlisting}
\end{tcolorbox}

\subsection{Subfigure Processing Prompts}
After obtaining refined subfigure-level captions, we further align each detected subfigure with its most relevant textual segment using a multimodal large language model~(MLLM). To enable precise alignment, we directly mark each detected subfigure in the image with a bounding box and a unique identifier (\emph{e.g.}, \textit{box\_1}, \textit{box\_2}), serving as a special visual prompt.
The text prompt shown below instructs the MLLM to carefully distinguish these detection box identifiers from any original panel labels (such as A, B, C) and to align each box with the most consistent fragment of the caption. The model is required to return the final alignment in a JSON format that explicitly records the mapping between each bounding box and its corresponding caption chunk. 

\label{suppsec:subfigure_processing}
\begin{tcolorbox}[
    colback=gray!5,
    colframe=gray!60,
    title={Subfigure Alignment Prompt},
    breakable
]
\begin{lstlisting}[
    basicstyle=\small\ttfamily,
    columns=fullflexible,
    breaklines=true,
    breakatwhitespace=true
]
System: System role (system)
You are an expert in scientific medical imaging text-image alignment. 

Please align the detection boxes in the uploaded object-detection visualization image with the given caption at a precise subfigure- subcaption level.

User role (user)
[Task Description]

Input:
1. Object-detection visualization image: The image contains multiple detection boxes. At the center of each detection box, there is a red rectangular marker, with the detection box identifier displayed inside a circle (in the format 'box_X', where X may be a letter or a number). If there are no detection boxes and corresponding identifiers in the image, it means no objects were detected.
2. An English caption describing the image content: "{caption_text}"

Important notes:
- The image may contain original subfigure labels (such as A, B, C, a, b, c, 1, 2, etc.); these are original labels of the figure.
- The 'box_Z', 'box_Y', 'box_X', etc. inside the red boxes are the detection box identifiers that we need to align; be sure to distinguish them from the original labels.
- The 'Z' in the detection box identifier 'box_Z' does not correspond to the original subfigure labels.

Task requirements:
    1. Carefully identify each detection box identifier inside the circles (in the 'box_X' format).
    2. Observe the specific position of each detection box in the image.
    3. Analyze the content described in the caption.
    4. Based on the actual location and content of each detection box, align it with the most relevant sub-description in the caption.

Alignment principles:
- Use the actual anatomical location and pointing direction of the detection box as the basis, rather than a superficial mapping between letters.
- If multiple detection boxes point to the same anatomical structure or described region, they may correspond to the same caption segment.
- If you cannot determine a matching relationship, mark it as "unknown".

Output format:
Only output the alignment result in JSON format. Do not output any non-JSON explanatory content:
[
    {"bbox_id": "detection box identifier", "caption_chunk": "corresponding original description segment"},
    {"bbox_id": "detection box identifier", "caption_chunk": "corresponding original description segment"},
    ...
]
\end{lstlisting}
\end{tcolorbox}

\section{Experiment Details}
\label{suppsec:experiment_details}

\subsection{Evaluation Prompts for Zero-shot Classification}
\label{suppsec:evaluation_prompts}
For zero-shot classification on PhenoBench, we employ a set of prompt templates inspired by CLIP~\cite{clip} and tailored to the medical imaging domain. Given a phenotype class name \texttt{[CLASS\_NAME]}, we instantiate multiple textual variants that describe the same concept from different clinical perspectives (\emph{e.g.}, diagnosis, clinical signs, abnormal findings). These templates form the text inputs to the text encoder for each class, and their embeddings are averaged to obtain a robust class representation. This strategy reduces sensitivity to specific phrasing, ensuring more robust evaluation. The full list of prompt templates is provided in the box below.

\begin{tcolorbox}[colback=gray!5,colframe=gray!60,title={Zero-Shot Classification Prompt Templates}]
    \small
    \textbf{Prompt Templates (for a class \texttt{[CLASS\_NAME]}):}
    \begin{itemize}
        \item \texttt{A medical image showing [CLASS\_NAME].}
        \item \texttt{Diagnosis of [CLASS\_NAME].}
        \item \texttt{Clinical signs of [CLASS\_NAME].}
        \item \texttt{Image from a patient with [CLASS\_NAME].}
        \item \texttt{This is a photo of [CLASS\_NAME].}
        \item \texttt{Findings consistent with [CLASS\_NAME].}
        \item \texttt{Evidence of [CLASS\_NAME].}
        \item \texttt{A case of [CLASS\_NAME].}
        \item \texttt{An example of [CLASS\_NAME].}
        \item \texttt{This image displays features of [CLASS\_NAME].}
        \item \texttt{Image confirms a diagnosis of [CLASS\_NAME].}
        \item \texttt{Abnormal findings suggesting [CLASS\_NAME].}
    \end{itemize}
\end{tcolorbox}

\subsection{Evaluation Baseline}
\label{suppsec:evaluation_baseline}
We consider two main categories of state-of-the-art vision–language models (VLMs) for comparison, \emph{i.e.}, {general-domain VLMs} and {biomedical VLMs}. 

\noindent The former category includes general-domain VLMs that are pre-trained on large-scale web image–text datasets, aiming to capture diverse and generic visual-semantic relationships:
\begin{itemize}
    \item \textbf{Open-CLIP}~\cite{openclip}: A widely-used open-source implementation of CLIP, trained on diverse image-text pairs from the web.
    \item \textbf{SigLIP2}~\cite{siglip2}: A CLIP-based model fine-tuned with additional supervision signals to enhance its performance on fine-grained visual-semantic tasks.
    \item \textbf{CoCa}~\cite{coca}: A contrastive captioning model that combines image-text alignment with generative captioning objectives, achieving strong performance on general vision-language benchmarks.
\end{itemize}

\noindent The latter category comprises biomedical VLMs that are specifically pre-trained on medical image–text datasets, focusing on domain-specific knowledge:
\begin{itemize}
    % 收集了1.6M的训练数据
    \item \textbf{PMC-CLIP}~\cite{pmc-clip}: A CLIP-based model trained on PubMed Central Open Access (PMC-OA)~\cite{pmcoa} image-caption pairs, designed to align medical images with their corresponding textual descriptions. The training dataset contains 1.6 million image-text pairs.
    % 收集了PMC-15M，并基于其进行了训练
    \item \textbf{BiomedCLIP}~\cite{biomedclip}: A biomedical adaptation of CLIP, trained on the PMC-15M dataset, which consists of 15 million image-text pairs collected from PubMed Central. 
    % 收集并发布了24M的PMC数据
    \item \textbf{BIOMEDICA}~\cite{biomedica}: A vision-language model trained on a large-scale dataset of 24 million image-text pairs collected and released from PMC-OA. 
\end{itemize}

\subsection{Evaluation Metrics}
Model performance is evaluated with metrics tailored to each setting. For zero-shot classification and linear probing, we report {classification accuracy (ACC)}. For cross-modal retrieval tasks, including the rare facial phenotype matching on {Face2Gene}~\cite{face2gene}, we use {Recall@K (R@K)} as the main evaluation metric, along with {Precision}, {Recall}, and {F1-Score} to capture fine-grained matching performance in clinically relevant retrieval scenarios.

\section{PhenoKG Construction Details}
\label{suppsec:phenoKG}
In this section, we describe our data curation pipeline in more detail.

\subsection{Knowledge Graph Initialization}
\label{suppsec:kg_construction_text}

\textbf{Phenotype knowledge collection.}
To establish a comprehensive phenotype knowledge graph, we collected a textual knowledge graph based on the Human Phenotype Ontology (HPO)~\cite{hpo}.
HPO is a standardized terminology system for phenotypes, widely used in clinical diagnosis and genetic disease research. 
We collected 19,703 phenotype terms from HPO, covering a wide range of clinical manifestations, including anatomical abnormalities and physiological dysfunctions. 
Additionally, we extracted 6,990 synonyms for each phenotype term to enhance the coverage of the terminology. 
Finally, we constructed a relationship network among phenotype terms, comprising 23,528 edges and 19,703 nodes that represent hierarchical relationships and correlations between phenotype terms.

\vspace{3pt} \noindent \textbf{Graph structure.}
The collected textual phenotype KG is structured as a directed acyclic graph (DAG), where nodes represent phenotype terms and edges represent relationships between terms.
Each node contains the following information: phenotype name, definition, and a list of synonyms.
Each directed edge represents a hierarchical relationship from a child phenotype to a parent phenotype.
We categorize the nodes in PhenoKG-Text into two types: group nodes and leaf nodes.
Group nodes represent broader phenotype categories with several child nodes.
Leaf nodes are terminal nodes in the knowledge tree, representing specific phenotype instances.

\vspace{3pt}

\subsection{Multimodal Data Collection and Processing}
\label{suppsec:phenopmc_collection}

After initializing the textual Phenotype KG, we augment each node with relevant image-caption pairs extracted from medical literature. 
As illustrated in Fig.~\ref{fig:pipeline_1}~(left), this process involves the following steps:

\begin{figure}[h]
    \centering
    \includegraphics[width=1\linewidth]{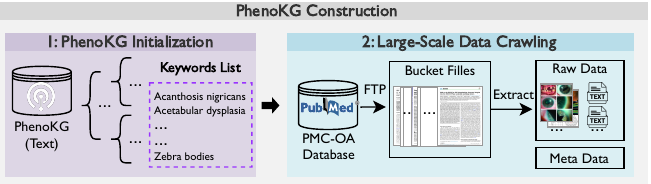}
    \caption{Text Initilization}
    \label{fig:pipeline_1}
\end{figure}

\vspace{3pt} \noindent \textbf{Keyword collection.}
To collect diverse medical image–caption pairs from medical literature, we first derived a list of phenotype-specific search keywords from the original textual Phenotype KG.
The primary goal was to ensure our search queries were semantically specific enough to retrieve precise clinical findings. To this end, we filtered out the broader, non-terminal nodes (e.g., ``Abnormality of the nervous system'') and exclusively retained the terminal nodes that represent the most specific phenotypes and have no child nodes in the hierarchy.
To enhance the robustness and coverage of our search, we then augmented this list by including all official synonyms for these terminal nodes. This process resulted in a final, curated list of 27,832 unique phenotype keywords, which formed the basis for our large-scale data collection.

\vspace{3pt} \noindent \textbf{Large-scale data crawling.}
Leveraging this keyword list, we processed the entire PubMed Central Open Access (PMCOA) full-text dataset to extract relevant image-caption pairs. Our automated pipeline parsed the JATS XML file of each article, iterating through all figures and their associated captions. For each figure, we used regular expressions to match our keyword list against its caption text. 
A pair was retained only if its caption contained an exact match for one of the phenotype keywords. In addition to the caption, we also extracted the corresponding paragraph from the main body of the text that cited the figure (e.g.,  ``...as seen in Figure 1, the fundus image reveals multiple microaneurysms.''), preserving rich contextual information.

\subsubsection{Multimodal Data Processing Details}
\label{sec:image_processing}

\begin{figure}[h]
    \centering
    \includegraphics[width=1\linewidth]{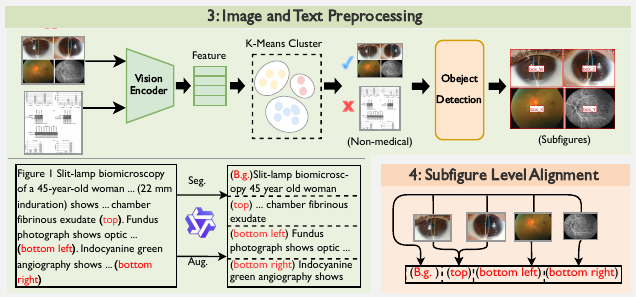}
    \caption{Multimodal Data  Processing}
    \label{fig:pipeline_2}
\end{figure}

\vspace{3pt} \noindent \textbf{Image processing.}
As shown in Fig.~\ref{fig:pipeline_2}, the crawled image-caption pairs often contain various types of image-level noise, \emph{i.e.}, compound figures and non-medical figures (\emph{e.g.}, charts and graphs). 
To address this issue, we first employed a cluster-based filtering procedure. More specifically, we adopt a pretrained visual encoder, DINOv3~\cite{dinov3}, to extract image embeddings and cluster them using K-Means into 20 main clusters, each further subdivided into 20 sub-clusters, resulting in a total of 400 clusters. 
We then manually inspected each cluster and retained only those marked as relevant to medical content. 
Afterwards, we processed compound figures by applying a pretrained subfigure detection model, DAB-DETR~\cite{dabdetr,openpmc-18m}, to all the remaining images. 
This model automatically splits compound figures into subfigures, extracting individual medical images. 
Finally, we applied the same cluster-based filtering procedure to the newly obtained subfigures to ensure data purity. 
As a result, we obtained 524{,}802 medical images linked with their corresponding texts.

\vspace{3pt} \noindent \textbf{Text processing.} After image processing, we proceeded to process the associated texts. 
As introduced earlier, for each crawled image, we preserved its corresponding caption as well as the main text paragraph that referred to the image. 
We then adopted a large language model (LLM), Qwen3~\cite{qwen3}, to extract key information corresponding to each image, cleaning irrelevant symbols and numerical artifacts. 
Furthermore, considering that some figures were separated into subfigures in the previous step, we additionally prompted the LLM to generate subfigure-level textual descriptions by splitting the 
% 详见附录
original descriptive text accordingly, leveraging a tailored prompt (detailed in Suppl. ~\ref{sec:text_splitting}).

\begin{figure}[H]
    \centering
    \includegraphics[width=0.9\linewidth]{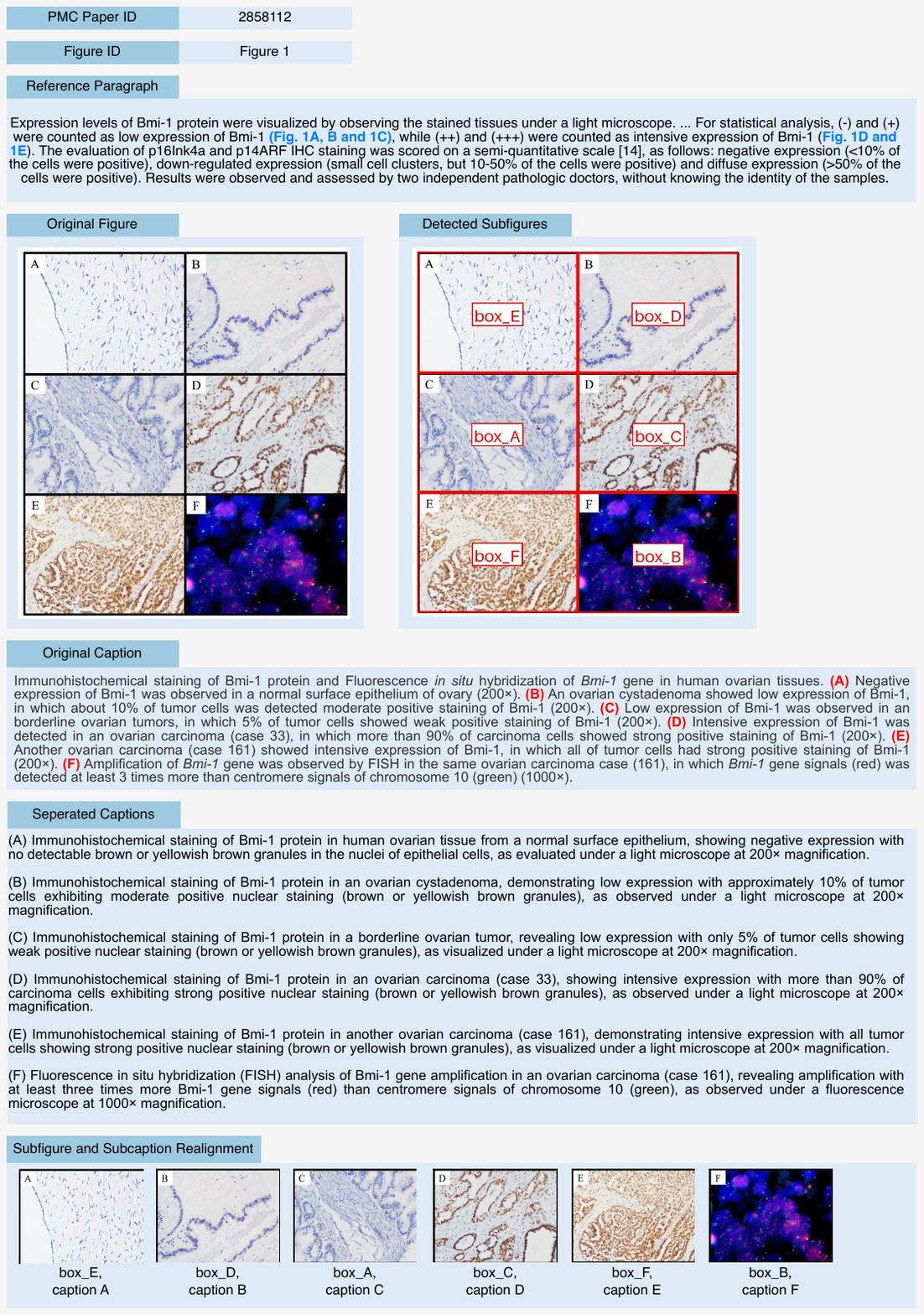}
    \caption{Case Visualization of Multimodal Data Processing }
    \label{suppfig:supp_vis}
\end{figure}

\vspace{3pt} \noindent \textbf{Sub-figure and text realignment.}
We realign the separated sub-figures and related split texts. 
To achieve this, we visualized the sub-figures detected in Step 3 within the original image to generate visual prompts:
we modify the original composite image by overlaying the detected bounding boxes. Crucially, as shown in the last part of Stage 3 in figure~\ref{fig:pipeline_2}, we embed a unique textual identifier (e.g., box\_W, box\_X) directly onto the image within each bounding box.
Using Qwen2.5VL~\cite{qwen2.5vl}, we input the visual prompt images along with the enhanced descriptive texts generated in Step 3 to achieve sub-figure level alignment.
\vspace{3pt} \noindent \textbf{Multimodal integration.} The final step in creating PhenoKG was to integrate the collected image-caption pairs into the original textual phenotype KG nodes. Considering each image-caption pair is linked to a corresponding phenotype node~(used as search keyword), it is straightforward to insert them into the original textual KG through keyword matching, resulting in the final PhenoKG, which is a large-scale, comprehensive multimodal phenotype-wise knowledge graph.

\section{PhenoLIP Details}

In this section, we introduce more details about PhenoLIP, including more strict loss formulations in PhenoLIP and implementation details.

\subsection{Implementation Details of Phenotype Knowledge Encoding}
\label{suppsec:kg_encoding}

We adopted the pretrained PubmedBERT~\cite{pubmedbert} as the initialization of the knowledge encoder. 
We trained the model using the AdamW optimizer with a learning rate of 1e-5, a batch size of 256, and a temperature parameter \(\tau_1\) set to 0.07. The training process lasted for 10 epochs on 8 NVIDIA A100 GPUs.

\subsection{More Method Details for Knowledge Distillation Loss }
\label{app:kd_loss}
As we illustrated in Section 4.2, we introduce an additional knowledge distillation branch in the process of vision-language pretraining. 
The pretrained phenotype knowledge encoder $\Phi_{\text{k}}$ is frozen and used as a teacher model. 
For each caption $c_i$ in the batch, the teacher produces a reference knowledge embedding 
$\mathbf{k}_i = \Phi_{\text{k}}(c_i)$, which serves as the ground-truth supervision 
for distilling phenotype-level knowledge into the text encoder. 
We then enforce consistency between the learnable text embedding ${t}_i$ 
and the frozen knowledge embedding ${k}_i = \Phi_\text{k}(c_i)$ through a text-knowledge contrastive loss. This loss encourages the learnable text encoder to be aligned within the phenotype knowledge embedding space, formulated as:
\begin{align}
\mathcal{L}_{\text{KD}} 
= & \frac{1}{B} \sum_{i=1}^{B}-\log 
\frac{\exp(\text{sim}(\mathbf{t}_i, \mathbf{k}_i)/\tau_3)}
{\sum_{j=1}^{B} \exp(\text{sim}(\mathbf{t}_i, \mathbf{k}_j)/\tau_3)}
\nonumber \\
& - \log 
\frac{\exp(\text{sim}(\mathbf{k}_i, \mathbf{t}_i)/\tau_3)}
{\sum_{j=1}^{B} \exp(\text{sim}(\mathbf{k}_i, \mathbf{t}_j)/\tau_3)},
\label{eq:kd_loss}
\end{align}
where $\tau_3$ is a temperature parameter. The final overall loss is a weighted combination of the vision-language contrastive loss and the knowledge distillation loss, averaged over all samples in a batch:
\begin{align}
    \mathcal{L}_{\text{VLP}} = \mathcal{L}_{\text{M}} + \alpha \mathcal{L}_{\text{KD}},
    \label{eq:total_loss}
\end{align}
where $\alpha$ is a hyperparameter. This dual-objective approach ensures that while the model learns to align images with their free-text captions, it is continuously guided by the structured, hierarchical knowledge from the PhenoKG, enhancing its ability to understand complex phenotypes.

\subsection{Implementation Details of Knowledge-enhanced VLP}
\label{suppsec:ke_vlp}
We adopted BiomedCLIP~\cite{biomedclip} as the initialization for the visual encoder, while the text encoder was initialized with the weights from the previously trained phenotype knowledge encoder.
The input images are resized to a resolution of 224x224. Text inputs are tokenized and embedded, with a maximum length set to 256 tokens. The training batch size is set to 256, with a learning rate of 1e-5, and employs 500 steps of linear warm-up followed by cosine annealing scheduling. The knowledge distillation weight hyperparameter \(\alpha\) is set to 0.3. The training process lasts for 10 epochs on 8 NVIDIA A100 GPUs.

{

\section{Comparison with Knowledge-Enhanced VLMs}
\label{suppsec:ke_comparison}

We provide a comprehensive comparison between PhenoLIP and existing knowledge-enhanced medical VLMs to demonstrate the effectiveness of our phenotype-centric approach across diverse medical imaging domains.

\subsection{Knowledge-Enhanced Baselines}
\label{suppsec:ke_baselines}

We compare PhenoLIP against several representative knowledge-enhanced approaches that incorporate medical knowledge into vision-language pretraining:

\begin{itemize}
    \item \textbf{DermLIP}~\cite{Derm1M}: A dermatology-focused VLM that constructs a dermatological ontology to refine textual descriptions. The ontology is primarily used for text enhancement without explicit graph structure guidance for visual feature learning.

    \item \textbf{KEEP}~\cite{keep}: A pathology-focused method that injects disease-level knowledge from pathology knowledge bases. It focuses primarily on textual knowledge integration rather than phenotype-level features.

    \item \textbf{MedKLIP/KAD}~\cite{medklip,kad}: Radiology-centric approaches that supervise training with knowledge-enhanced triples extracted from expert-written radiology reports. These methods leverage structured report annotations for knowledge injection.
\end{itemize}

\vspace{3pt}\noindent\textbf{Methodological distinctions:}
PhenoLIP differs from these approaches in two key aspects. \textit{First}, in terms of application scope, unlike domain-specific models (DermLIP for dermatology, KEP/KEEP for pathology, MedKLIP/KAD for radiology), PhenoLIP centers on phenotypes-observable traits that inherently span across modalities and anatomical systems. This phenotype-centric design allows the model to operate beyond traditional medical imaging domains, encompassing diverse visual manifestations including facial features, skin lesions, and radiological findings. \textit{Second}, methodologically, while DermLIP uses ontologies primarily for text refinement and MedKLIP/KAD rely on report-extracted triples, PhenoLIP distills the phenotype-ontology structure into a dedicated knowledge encoder to guide vision-language pretraining. This provides finer phenotype-level guidance for medical imaging analysis, yielding stronger generalization especially in data-scarce scenarios.

\subsection{Cross-domain performance comparison}
\label{suppsec:ke_results}

Table~\ref{supptab:ke_comparison} presents a comprehensive comparison across seven diverse medical imaging domains: Radiology, Dermatology, Pathology, Hematology, Histology, Ophthalmology, and Phenotype recognition.

\begin{table*}[h]
    \centering
    \small
    \setlength{\tabcolsep}{4pt}
    \begin{tabular}{lccccccc|c}
    \toprule
    \textbf{Method} & \textbf{Radiology} & \textbf{Dermatology} & \textbf{Pathology} & \textbf{Hematology} & \textbf{Histology} & \textbf{Ophthalmology} & \textbf{Phenotype} & \textbf{Average} \\
    \midrule
    DermLIP & 31.6 & \textbf{62.0} & 19.1 & 5.6 & 6.6 & \underline{25.2} & \underline{2.4} & 21.8 \\
    KEEP & 14.0 & 23.6 & \textbf{91.5} & 8.7 & \underline{8.3} & 21.5 & 1.7 & \underline{24.2} \\
    KAD & \underline{41.4} & 31.3 & 22.1 & \underline{9.1} & 3.7 & 22.6 & 1.7 & 18.8 \\
    \midrule
    \textbf{PhenoLIP (Ours)} & \textbf{49.0} & \underline{55.1} & \underline{58.3} & \textbf{23.2} & \textbf{7.9} & \textbf{51.6} & \textbf{10.8} & \textbf{36.6} \\
    \bottomrule
    \end{tabular}
    \caption{Comprehensive comparison with knowledge-enhanced and domain-specific VLMs across diverse medical imaging domains. PhenoLIP achieves the best average performance (36.6\%), demonstrating superior generalization across modalities. While domain-specific specialists excel in their target domains (DermLIP: 62.0\% in Dermatology, KEEP: 91.5\% in Pathology), they show limited transferability to other domains. PhenoLIP achieves the best performance in 5 out of 7 domains, validating the effectiveness of phenotype-centric knowledge integration.}
    \label{supptab:ke_comparison}
\end{table*}

\noindent\textbf{Performance comparison.}
PhenoLIP achieves the best performance in 5 out of 7 domains, demonstrating its ability to generalize across diverse medical imaging modalities without domain-specific fine-tuning. While domain specialists such as DermLIP (62.0\% in Dermatology) and KEEP (91.5\% in Pathology) achieve strong results in their target domains, they show limited transferability and often perform poorly outside their specialization. Notably, PhenoLIP achieves the highest performance on phenotype recognition (10.8\%), significantly outperforming all knowledge-enhanced baselines, which validates the effectiveness of our phenotype-centric knowledge integration strategy.
}
\section{Ablation study details}
\label{suppsec:ablation}
{

\subsection{Knowledge graph component ablation}
\label{suppsec:kg_ablation}

To validate the contribution of different knowledge graph components in the knowledge encoder, we conducted an ablation study by systematically removing definitions, synonyms, and relational structure from PhenoKG. The results are shown in Table~\ref{supptab:kg_ablation}.

\begin{table}[h]
    \footnotesize
    \centering
    \setlength{\tabcolsep}{3pt}
    \begin{tabular}{lccccc}
    \toprule
    \textbf{Setting} & \textbf{I2T R@10} & \textbf{I2T R@50} & \textbf{T2I R@10} & \textbf{T2I R@50} & \textbf{Avg.($\Delta$)} \\
    \midrule
    \rowcolor{gray!20} Full Model & {63.88} & {81.92} & {66.61} & {87.68} & {75.02}( -- ) \\
    \midrule
    $w/o$ Definitions  & 62.64 & 81.91 & 66.17 & 87.23 & 74.49\,(\textcolor{OliveGreen}{\textbf{$\downarrow$0.53}}) \\
    $w/o$ Synonyms & 63.16 & {81.92} & {66.61} & 87.39 & 74.77\,(\textcolor{OliveGreen}{{$\downarrow$0.25}}) \\
    $w/o$ Relations  & 62.30 & 81.89 & 66.37 & 87.59 & 74.54\,(\textcolor{OliveGreen}{\underline{$\downarrow$0.48}}) \\
    \bottomrule
    \end{tabular}
    \caption{Ablation study on knowledge graph components. We evaluate the contribution of each component by removing them from the knowledge encoder training. Results show that definitions cause the largest performance drop (0.53\%), followed closely by relational structure (0.48\%), demonstrating that both semantic definitions and graph topology are critical for effective phenotype knowledge encoding. Synonyms contribute less (0.25\%) but still improve model robustness by providing lexical variations.}
    \label{supptab:kg_ablation}
\end{table}

\noindent\textbf{Component contribution analysis.}
Removing definitions results in the largest performance drop (0.53\%), indicating that semantic understanding of phenotype concepts is essential for effective knowledge encoding. Removing relational structure causes a drop of 0.48\%, demonstrating that capturing hierarchical dependencies through graph topology contributes nearly as much as semantic definitions.  While synonyms contribute less individually (0.25\%), they still improve performance by providing lexical variations and increasing the coverage of phenotype terminology.
This ablation study confirms that our knowledge encoder effectively captures both semantic and structural information from PhenoKG, and that both aspects contribute significantly to the model's phenotype understanding capabilities.
}
\subsection{Training Component Ablation}
\label{suppsec:training_ablation}

We conduct a comprehensive ablation study to evaluate the impact of different training components on model performance. The results are shown in Table~\ref{supptab:ablation}.

% ablation
\begin{table}[h]
    \footnotesize
    \centering
    \setlength{\tabcolsep}{3pt}
    \begin{tabular}{cccccccccc}
    \toprule
    \multicolumn{2}{c}{\textbf{Encoder}} & \textbf{Know.} & \textbf{Data} & \multicolumn{2}{c}{\textbf{I2T}} & \multicolumn{2}{c}{\textbf{T2I}} & \textbf{Cls.} \\
    \cmidrule(lr){1-2} \cmidrule(lr){3-3} \cmidrule(lr){4-4} \cmidrule(lr){5-6} \cmidrule(lr){7-8} \cmidrule(lr){9-9}
    \textbf{Vision} & \textbf{Text} & \textbf{Distill.} & \textbf{Cur.} & \textbf{R@10} & \textbf{R@50} & \textbf{R@10} & \textbf{R@50} & \textbf{Acc} \\ 
    \midrule
    Scratch & KB & & & 28.15 & 45.33 & 31.04 & 49.81 & 2.13 \\
    CLIP & KB & & & 47.20 & 68.91 & 51.72 & 73.05 & 6.44 \\ 
    BCLIP & KB & & & 50.11 & 71.22 & 53.68 & 75.90 & 7.02 \\ 
    CLIP & PMB & & & 49.53 & 70.88 & 54.88 & 78.14 & 6.95 \\
    BCLIP & PMB & & & 53.42 & 74.10 & 58.03 & 81.25 & 7.81 \\ 
    CLIP & PMB & \checkmark & & 58.91 & 78.54 & 62.19 & 84.33 & 9.57 \\
    CLIP & PMB & \checkmark & \checkmark & 60.13 & 79.62 & 63.55 & 85.18 & 9.98 \\
    \rowcolor{green!15}
    BCLIP & PMB & \checkmark & \checkmark & \textbf{63.30} & \textbf{81.92} & \textbf{66.61} & \textbf{87.68} & \textbf{10.76} \\
    \bottomrule
    \end{tabular}
    \vspace{-2mm}
    \caption{Ablation study on training components. We systematically evaluate the impact of encoder initializations, knowledge distillation, and data curation strategies on performance. BCLIP denotes BiomedCLIP's vision encoder, PMB denotes PubmedBert, and KB denotes a generic knowledge-base text encoder. The full model combining biomedical encoders, knowledge distillation, and data curation achieves the best results across all tasks.}
    \label{supptab:ablation}
\end{table}

To demonstrate the impact of each component of our method, we conduct a series of ablation studies on (\textit{i}) different encoder initializations, (\textit{ii}) our proposed knowledge distillation loss, and (\textit{iii}) data curation strategies with the cross-modal retrieval task on PhenoBench. The results are detailed in Table~\ref{supptab:ablation}.

\vspace{3pt} \noindent \textbf{Effect of encoder initialization.}
We first examine the impact of pre-trained weights on both the vision and text encoders. As shown in Table~\ref{supptab:ablation}, a model trained entirely from scratch performs poorly, achieving only 28.15\% on I2T R@10 and 2.13\% on classification accuracy. 
Initializing the encoders with general-domain CLIP and a generic knowledge-base text encoder(KB) leads to substantial improvements, raising I2T R@10 to 47.20\% and classification accuracy to 6.44\%. Furthermore, replacing the generic vision encoder with the biomedical-domain BiomedCLIP (BCLIP) improves I2T R@10 to 50.11\%. Similarly, substituting the generic text encoder with the PubmedBert \textit{PMB} enhances T2I R@10 from 51.72\% to 54.88\% when paired with CLIP. Combining both \textit{BCLIP} and \textit{PMB} establishes a strong baseline, yielding 53.42\% on I2T R@10 and 58.03\% on T2I R@10. These results confirm that leveraging encoders pre-trained on biomedical data provides a markedly stronger initialization, leading to more effective cross-modal alignment for our phenotype-centric task.

\vspace{3pt} \noindent \textbf{Effect of knowledge distillation loss.}
To validate our core contribution, we analyze the effect of the knowledge distillation loss ($\mathcal{L}_{\text{KD}}$). We compare the ``\textit{CLIP} + \textit{PMB}'' configuration in two variants: with and without the incorporation of $\mathcal{L}_{\text{KD}}$. The results reveal a substantial performance gain that adding $\mathcal{L}_{\text{KD}}$ to the ``\textit{CLIP} + \textit{PMB}'' configuration raises I2T R@10 from 49.53\% to 58.91\% and classification accuracy from 6.95\% to 9.57\%. Most critically, when applied to our best encoder setup (``\textit{BCLIP} + \textit{PMB}''), knowledge distillation emerges as a significant source of improvement.

\vspace{3pt} \noindent \textbf{Effect of data curation.}
Finally, we analyze the impact of our data curation strategies, as illustrated in Section~\ref{para:txt_processiing}. We compare the performance of models trained with and without these curations. Adding data curation to the ``CLIP + PMB + $\mathcal{L}_\text{KD}$'' model provides a consistent improvement, raising I2T R@10 from 58.91\%  to 60.13\%  and T2I R@10 from 62.19\%  to 63.55\%. This trend continues in our final model configuration. This demonstrates that our curation techniques effectively enhance model robustness and contribute to better generalization, leading to the final state-of-the-art performance.

%% file: main.bib
@String(ICCV= {Int. Conf. Comput. Vis.})

@String(ICLR = {Int. Conf. Learn. Represent.})

@String(AAAI = {AAAI})

@String(ICCV  = {ICCV})

@String(ICLR  = {ICLR})

@article{qiu2025evolving,
  title={Evolving Diagnostic Agents in a Virtual Clinical Environment},
  author={Qiu, Pengcheng and Wu, Chaoyi and Liu, Junwei and Zheng, Qiaoyu and Liao, Yusheng and Wang, Haowen and Yue, Yun and Fan, Qianrui and Zhen, Shuai and Wang, Jian and others},
  journal={arXiv preprint arXiv:2510.24654},
  year={2025}
}

@inproceedings{clip,
  title={Learning transferable visual models from natural language supervision},
  author={Radford, Alec and Kim, Jong Wook and Hallacy, Chris and Ramesh, Aditya and Goh, Gabriel and Agarwal, Sandhini and Sastry, Girish and Askell, Amanda and Mishkin, Pamela and Clark, Jack and others},
  booktitle={International conference on machine learning},
  pages={8748--8763},
  year={2021},
  organization={PmLR}
}

@article{hpo,
  title={The Human Phenotype Ontology in 2024: phenotypes around the world.},
  author={Talapova, P and Gargano, MA and Matentzoglu, N and Coleman, B and Addo-Lartey, EB and Anagnostopoulos, AV and Anderton, J and Avillach, P},
  year={2023},
  publisher={Nucleic Acids Research, Oxford University Press}
}

@inproceedings{biomedica,
  title={Biomedica: An open biomedical image-caption archive, dataset, and vision-language models derived from scientific literature},
  author={Lozano, Alejandro and Sun, Min Woo and Burgess, James and Chen, Liangyu and Nirschl, Jeffrey J and Gu, Jeffrey and Lopez, Ivan and Aklilu, Josiah and Rau, Anita and Katzer, Austin Wolfgang and others},
  booktitle={Proceedings of the Computer Vision and Pattern Recognition Conference},
  pages={19724--19735},
  year={2025}
}

@InProceedings{Derm1M,
    author    = {Yan, Siyuan and Hu, Ming and Jiang, Yiwen and Li, Xieji and Fei, Hao and Tschandl, Philipp and Kittler, Harald and Ge, Zongyuan},
    title     = {Derm1M: A Million-scale Vision-Language Dataset Aligned with Clinical Ontology Knowledge for Dermatology},
    booktitle = {Proceedings of the IEEE/CVF International Conference on Computer Vision (ICCV)},
    month     = {October},
    year      = {2025},
    pages     = {12681-12690}
}

@article{quilt1m,
  title={Quilt-1m: One million image-text pairs for histopathology},
  author={Ikezogwo, Wisdom and Seyfioglu, Saygin and Ghezloo, Fatemeh and Geva, Dylan and Sheikh Mohammed, Fatwir and Anand, Pavan Kumar and Krishna, Ranjay and Shapiro, Linda},
  journal={Advances in neural information processing systems},
  volume={36},
  pages={37995--38017},
  year={2023}
}

@article{biomedclip,
  title={A multimodal biomedical foundation model trained from fifteen million image--text pairs},
  author={Zhang, Sheng and Xu, Yanbo and Usuyama, Naoto and Xu, Hanwen and Bagga, Jaspreet and Tinn, Robert and Preston, Sam and Rao, Rajesh and Wei, Mu and Valluri, Naveen and others},
  journal={NEJM AI},
  volume={2},
  number={1},
  pages={AIoa2400640},
  year={2025},
  publisher={Massachusetts Medical Society}
}

@InProceedings{medklip,
    author    = {Wu, Chaoyi and Zhang, Xiaoman and Zhang, Ya and Wang, Yanfeng and Xie, Weidi},
    title     = {MedKLIP: Medical Knowledge Enhanced Language-Image Pre-Training for X-ray Diagnosis},
    booktitle = {Proceedings of the IEEE/CVF International Conference on Computer Vision (ICCV)},
    month     = {October},
    year      = {2023},
    pages     = {21372-21383}
}

@article{kad,
  title={Knowledge-enhanced visual-language pre-training on chest radiology images},
  author={Zhang, Xiaoman and Wu, Chaoyi and Zhang, Ya and Xie, Weidi and Wang, Yanfeng},
  journal={Nature Communications},
  volume={14},
  number={1},
  pages={4542},
  year={2023},
  publisher={Nature Publishing Group UK London}
}

@article{scikit,
  title={Scikit-learn: Machine learning in Python},
  author={Pedregosa, Fabian and Varoquaux, Ga{\"e}l and Gramfort, Alexandre and Michel, Vincent and Thirion, Bertrand and Grisel, Olivier and Blondel, Mathieu and Prettenhofer, Peter and Weiss, Ron and Dubourg, Vincent and others},
  journal={the Journal of machine Learning research},
  volume={12},
  pages={2825--2830},
  year={2011},
  publisher={JMLR. org}
}

@inproceedings{pmc-clip,
  title={Pmc-clip: Contrastive language-image pre-training using biomedical documents},
  author={Lin, Weixiong and Zhao, Ziheng and Zhang, Xiaoman and Wu, Chaoyi and Zhang, Ya and Wang, Yanfeng and Xie, Weidi},
  booktitle={International Conference on Medical Image Computing and Computer-Assisted Intervention},
  pages={525--536},
  year={2023},
  organization={Springer}
}

@inproceedings{cxr8,
  title={Chestx-ray8: Hospital-scale chest x-ray database and benchmarks on weakly-supervised classification and localization of common thorax diseases},
  author={Wang, Xiaosong and Peng, Yifan and Lu, Le and Lu, Zhiyong and Bagheri, Mohammadhadi and Summers, Ronald M},
  booktitle={Proceedings of the IEEE conference on computer vision and pattern recognition},
  pages={2097--2106},
  year={2017}
}

@article{ikraph,
  title={A comprehensive large-scale biomedical knowledge graph for AI-powered data-driven biomedical research},
  author={Zhang, Yuan and Sui, Xin and Pan, Feng and Yu, Kaixian and Li, Keqiao and Tian, Shubo and Erdengasileng, Arslan and Han, Qing and Wang, Wanjing and Wang, Jianan and others},
  journal={Nature Machine Intelligence},
  pages={1--13},
  year={2025},
  publisher={Nature Publishing Group UK London}
}

@article{primekg,
  title={Building a knowledge graph to enable precision medicine},
  author={Chandak, Payal and Huang, Kexin and Zitnik, Marinka},
  journal={Scientific Data},
  volume={10},
  number={1},
  pages={67},
  year={2023},
  publisher={Nature Publishing Group UK London}
}

@inproceedings{kep,
  title={Knowledge-enhanced visual-language pretraining for computational pathology},
  author={Zhou, Xiao and Zhang, Xiaoman and Wu, Chaoyi and Zhang, Ya and Xie, Weidi and Wang, Yanfeng},
  booktitle={European Conference on Computer Vision},
  pages={345--362},
  year={2024},
  organization={Springer}
}

@inproceedings{dabdetr,
  author       = {Shilong Liu and
                  Feng Li and
                  Hao Zhang and
                  Xiao Yang and
                  Xianbiao Qi and
                  Hang Su and
                  Jun Zhu and
                  Lei Zhang},
  title        = {{DAB-DETR:} Dynamic Anchor Boxes are Better Queries for {DETR}},
  booktitle    = {The Tenth International Conference on Learning Representations, {ICLR}
                  2022, Virtual Event, April 25-29, 2022},
  publisher    = {OpenReview.net},
  year         = {2022},
  url          = {https://openreview.net/forum?id=oMI9PjOb9Jl},
  bibsource    = {dblp computer science bibliography, https://dblp.org}
}

@article{openpmc-18m,
  title={Open-PMC-18M: A High-Fidelity Large Scale Medical Dataset for Multimodal Representation Learning},
  author={Baghbanzadeh, Negin and Ashkezari, Sajad and Dolatabadi, Elham and Afkanpour, Arash},
  journal={arXiv preprint arXiv:2506.02738},
  year={2025}
}

@article{dinov3,
  title={Dinov3},
  author={Sim{\'e}oni, Oriane and Vo, Huy V and Seitzer, Maximilian and Baldassarre, Federico and Oquab, Maxime and Jose, Cijo and Khalidov, Vasil and Szafraniec, Marc and Yi, Seungeun and Ramamonjisoa, Micha{\"e}l and others},
  journal={arXiv preprint arXiv:2508.10104},
  year={2025}
}

@article{qwen3,
  title={Qwen3 technical report},
  author={Yang, An and Li, Anfeng and Yang, Baosong and Zhang, Beichen and Hui, Binyuan and Zheng, Bo and Yu, Bowen and Gao, Chang and Huang, Chengen and Lv, Chenxu and others},
  journal={arXiv preprint arXiv:2505.09388},
  year={2025}
}

@article{qwen2.5vl,
  title={Qwen2. 5-vl technical report},
  author={Bai, Shuai and Chen, Keqin and Liu, Xuejing and Wang, Jialin and Ge, Wenbin and Song, Sibo and Dang, Kai and Wang, Peng and Wang, Shijie and Tang, Jun and others},
  journal={arXiv preprint arXiv:2502.13923},
  year={2025}
}

@article{ham10000,
  title={The HAM10000 dataset, a large collection of multi-source dermatoscopic images of common pigmented skin lesions},
  author={Tschandl, Philipp and Rosendahl, Cliff and Kittler, Harald},
  journal={Scientific data},
  volume={5},
  number={1},
  pages={1--9},
  year={2018},
  publisher={Nature Publishing Group}
}

@misc{siglip2,
      title={SigLIP 2: Multilingual Vision-Language Encoders with Improved Semantic Understanding, Localization, and Dense Features}, 
      author={Michael Tschannen and Alexey Gritsenko and Xiao Wang and Muhammad Ferjad Naeem and Ibrahim Alabdulmohsin and Nikhil Parthasarathy and Talfan Evans and Lucas Beyer and Ye Xia and Basil Mustafa and Olivier Hénaff and Jeremiah Harmsen and Andreas Steiner and Xiaohua Zhai},
      year={2025},
      eprint={2502.14786},
      archivePrefix={arXiv},
      primaryClass={cs.CV},
      url={https://arxiv.org/abs/2502.14786}, 
}

@article{coca,
  author       = {Jiahui Yu and
                  Zirui Wang and
                  Vijay Vasudevan and
                  Legg Yeung and
                  Mojtaba Seyedhosseini and
                  Yonghui Wu},
  title        = {CoCa: Contrastive Captioners are Image-Text Foundation Models},
  journal      = {Trans. Mach. Learn. Res.},
  volume       = {2022},
  year         = {2022},
  url          = {https://openreview.net/forum?id=Ee277P3AYC},
  bibsource    = {dblp computer science bibliography, https://dblp.org}
}

@article{lc25000,
  title={Lung and colon cancer histopathological image dataset (lc25000)},
  author={Borkowski, Andrew A and Bui, Marilyn M and Thomas, L Brannon and Wilson, Catherine P and DeLand, Lauren A and Mastorides, Stephen M},
  journal={arXiv preprint arXiv:1912.12142},
  year={2019}
}

@article{radfm,
  title={Towards generalist foundation model for radiology by leveraging web-scale 2d\&3d medical data},
  author={Wu, Chaoyi and Zhang, Xiaoman and Zhang, Ya and Hui, Hui and Wang, Yanfeng and Xie, Weidi},
  journal={Nature Communications},
  volume={16},
  number={1},
  pages={7866},
  year={2025},
  publisher={Nature Publishing Group UK London}
}

@article{llavamed,
  title={Llava-med: Training a large language-and-vision assistant for biomedicine in one day},
  author={Li, Chunyuan and Wong, Cliff and Zhang, Sheng and Usuyama, Naoto and Liu, Haotian and Yang, Jianwei and Naumann, Tristan and Poon, Hoifung and Gao, Jianfeng},
  journal={Advances in Neural Information Processing Systems},
  volume={36},
  pages={28541--28564},
  year={2023}
}

@article{laion,
  title={Laion-5b: An open large-scale dataset for training next generation image-text models},
  author={Schuhmann, Christoph and Beaumont, Romain and Vencu, Richard and Gordon, Cade and Wightman, Ross and Cherti, Mehdi and Coombes, Theo and Katta, Aarush and Mullis, Clayton and Wortsman, Mitchell and others},
  journal={Advances in neural information processing systems},
  volume={35},
  pages={25278--25294},
  year={2022}
}

@inproceedings{chexpert,
  title={Chexpert: A large chest radiograph dataset with uncertainty labels and expert comparison},
  author={Irvin, Jeremy and Rajpurkar, Pranav and Ko, Michael and Yu, Yifan and Ciurea-Ilcus, Silviana and Chute, Chris and Marklund, Henrik and Haghgoo, Behzad and Ball, Robyn and Shpanskaya, Katie and others},
  booktitle={Proceedings of the AAAI conference on artificial intelligence},
  volume={33},
  number={01},
  pages={590--597},
  year={2019}
}

@article{medmnistv2,
  title={Medmnist v2-a large-scale lightweight benchmark for 2d and 3d biomedical image classification},
  author={Yang, Jiancheng and Shi, Rui and Wei, Donglai and Liu, Zequan and Zhao, Lin and Ke, Bilian and Pfister, Hanspeter and Ni, Bingbing},
  journal={Scientific Data},
  volume={10},
  number={1},
  pages={41},
  year={2023},
  publisher={Nature Publishing Group UK London}
}

@article{face2gene,
  title={Identifying facial phenotypes of genetic disorders using deep learning},
  author={Gurovich, Yaron and Hanani, Yair and Bar, Omri and Nadav, Guy and Fleischer, Nicole and Gelbman, Dekel and Basel-Salmon, Lina and Krawitz, Peter M and Kamphausen, Susanne B and Zenker, Martin and others},
  journal={Nature medicine},
  volume={25},
  number={1},
  pages={60--64},
  year={2019},
  publisher={Nature Publishing Group US New York}
}

@article{umls,
  title={The unified medical language system (UMLS): integrating biomedical terminology},
  author={Bodenreider, Olivier},
  journal={Nucleic acids research},
  volume={32},
  number={suppl\_1},
  pages={D267--D270},
  year={2004},
  publisher={Oxford University Press}
}

@article{orphanet,
  title={Orphanet: a European database for rare diseases},
  author={Weinreich, Steffanie S and Mangon, R and Sikkens, JJ and Teeuw, ME En and Cornel, MC},
  journal={Nederlands tijdschrift voor geneeskunde},
  volume={152},
  number={9},
  pages={518--519},
  year={2008}
}

@article{omim,
  title={OMIM. org: Online Mendelian Inheritance in Man (OMIM{\textregistered}), an online catalog of human genes and genetic disorders},
  author={Amberger, Joanna S and Bocchini, Carol A and Schiettecatte, Fran{\c{c}}ois and Scott, Alan F and Hamosh, Ada},
  journal={Nucleic acids research},
  volume={43},
  number={D1},
  pages={D789--D798},
  year={2015},
  publisher={Oxford University Press}
}

@misc{chestxreasoner,
      title={ChestX-Reasoner: Advancing Radiology Foundation Models with Reasoning through Step-by-Step Verification}, 
      author={Ziqing Fan and Cheng Liang and Chaoyi Wu and Ya Zhang and Yanfeng Wang and Weidi Xie},
      year={2025},
      eprint={2504.20930},
      archivePrefix={arXiv},
      primaryClass={cs.AI},
      url={https://arxiv.org/abs/2504.20930}, 
}

@misc{pmcoa,
  title={PubMed Central: The GenBank of the published literature},
  author={Roberts, Richard J},
  journal={Proceedings of the National Academy of Sciences},
  volume={98},
  number={2},
  pages={381--382},
  year={2001},
  publisher={The National Academy of Sciences}
}

@article{pubmedbert,
  title={Domain-specific language model pretraining for biomedical natural language processing},
  author={Gu, Yu and Tinn, Robert and Cheng, Hao and Lucas, Michael and Usuyama, Naoto and Liu, Xiaodong and Naumann, Tristan and Gao, Jianfeng and Poon, Hoifung},
  journal={ACM Transactions on Computing for Healthcare (HEALTH)},
  volume={3},
  number={1},
  pages={1--23},
  year={2021},
  publisher={ACM New York, NY}
}

@inproceedings{openclip,
  title={Reproducible scaling laws for contrastive language-image learning},
  author={Cherti, Mehdi and Beaumont, Romain and Wightman, Ross and Wortsman, Mitchell and Ilharco, Gabriel and Gordon, Cade and Schuhmann, Christoph and Schmidt, Ludwig and Jitsev, Jenia},
  booktitle={Proceedings of the IEEE/CVF Conference on Computer Vision and Pattern Recognition},
  pages={2818--2829},
  year={2023}
}

@misc{keep,
      title={Knowledge-enhanced Pretraining for Vision-language Pathology Foundation Model on Cancer Diagnosis}, 
      author={Xiao Zhou and Luoyi Sun and Dexuan He and Wenbin Guan and Ge Wang and Ruifen Wang and Lifeng Wang and Xiaojun Yuan and Xin Sun and Ya Zhang and Kun Sun and Yanfeng Wang and Weidi Xie},
      year={2026},
      eprint={2412.13126},
      archivePrefix={arXiv},
      primaryClass={eess.IV},
      url={https://arxiv.org/abs/2412.13126}, 
}
